\documentclass[10pt,twocolumn,letterpaper]{article}

\usepackage{iccv}

\usepackage{times}
\usepackage{relsize}
\usepackage[T1]{fontenc}
\usepackage[latin1]{inputenc}
\usepackage[english]{babel}
\usepackage{anyfontsize}

\usepackage{epsfig}
\usepackage{graphicx}
\usepackage{wrapfig}
\usepackage[belowskip=0pt,aboveskip=3pt,font=small]{caption}
\usepackage[belowskip=0pt,aboveskip=0pt,font=small]{subcaption}
\usepackage{amsmath, amsthm, amssymb, mathabx}
\usepackage{textcomp}
\usepackage{stmaryrd}
\SetSymbolFont{stmry}{bold}{U}{stmry}{m}{n}
\usepackage{upgreek}
\usepackage{bm}
\usepackage{cases}
\usepackage{mathtools}
\usepackage{arydshln}
\usepackage{multirow}

\usepackage{cite}
\usepackage[pagebackref=true,breaklinks=true,letterpaper=true,colorlinks,bookmarks=false]{hyperref}
\hypersetup{colorlinks,linkcolor={green},citecolor={green},urlcolor={red}}  

\usepackage{bibspacing}
\usepackage{algorithm}
\usepackage{algpseudocode}

\usepackage{multirow}
\usepackage{rotating}
\usepackage{booktabs}

\usepackage{enumitem}
\usepackage[olditem,oldenum]{paralist}

\usepackage{alltt}
\usepackage{listings}

\usepackage{mysymbols}

\usepackage{url}
\usepackage{xspace}
\usepackage{setspace}
\usepackage{comment}
\usepackage{color}
\usepackage[dvipsnames]{xcolor}
\usepackage{afterpage}
\usepackage{pdfpages}
\usepackage{framed}
\usepackage{fancybox}
\usepackage{cuted}
\usepackage{caption}
\usepackage{tabularx}

\usepackage{quoting}
\quotingsetup{vskip=0pt}
\usepackage{epigraph}
\usepackage[breaklinks=true,bookmarks=false]{hyperref}
\usepackage{diagbox}

\usepackage{dblfloatfix}

\usepackage{inconsolata}

\usepackage[pagebackref=true,breaklinks=true,letterpaper=true,colorlinks,bookmarks=false]{hyperref}

\iccvfinalcopy 

\def\iccvPaperID{31} 

\ificcvfinal\fi

\newcommand{\ttbf}[1]{\texttt{\textbf{#1}}}
\newcommand{\nocaps}[0]{\ttbf{nocaps}\xspace}
\newcommand{\nocapsin}[0]{\ttbf{in-domain}\xspace}
\newcommand{\nocapsnear}[0]{\ttbf{near-domain}\xspace}
\newcommand{\nocapsout}[0]{\ttbf{out-of-domain}\xspace}

\newcommand{\coco}[0]{COCO\xspace}
\newcommand{\openimages}[0]{Open Images\xspace}
\newcommand{\visualgenome}[0]{Visual Genome\xspace}
\newcommand{\updown}[0]{UpDown\xspace}

\newcommand{\nbt}[1]{\textcolor{RubineRed}{\textbf{#1}}}
\newcommand{\cbs}[1]{\textcolor{NavyBlue}{\textbf{#1}}}

\newcommand{\indomain}[1]{\textcolor{Orange}{\textbf{#1}}}
\newcommand{\outdomain}[1]{\textcolor{RoyalBlue}{\textbf{#1}}}

\newcolumntype{Y}{>{\centering\arraybackslash}X}

\newcommand{\oidet}[0]{\ttbf{OI detector}\xspace}
\newcommand{\vgdet}[0]{\ttbf{VG detector}\xspace}

\belowdisplayskip \abovedisplayskip

\newlength{\abstractReduceTop}
\newlength{\abstractReduceBot}

\newlength{\sectionReduceTop}
\newlength{\sectionReduceBot}

\newlength{\subsectionReduceTop}
\newlength{\subsectionReduceBot}

\newlength{\subsubsectionReduceTop}
\newlength{\subsubsectionReduceBot}

\newlength{\captionReduceTop}
\newlength{\captionReduceBot}

\newlength{\eqnReduceTop}
\newlength{\eqnReduceBot}

\newlength{\horSkip}
\newlength{\verSkip}

\newlength{\figureHeight}
\makeatletter
\@namedef{ver@everyshi.sty}{}
\makeatother
\usepackage{tikz}
\usetikzlibrary{automata, positioning, arrows}
\tikzset{
->, 
node distance=2cm, 
initial text=$ $,
}

\usepackage{multicol,caption}

\usepackage{mathtools,eqparbox}

\newenvironment{Figure}
  {\par\medskip\noindent\minipage{\linewidth}}
  {\endminipage\par\medskip}

\begin{document}

\title{\nocaps{}: novel object captioning at scale}

\author{
    Harsh Agrawal$^{* 1}$\\
    Mark Johnson$^{2}$
    \and
    Karan Desai$^{* 1,4}$\\
    Dhruv Batra$^{1,3}$
    \and
    Yufei Wang$^{2}$\\
    Devi Parikh$^{1,3}$
    \and
    Xinlei Chen$^3$\\
    Stefan Lee$^{1,5}$
    \and
    Rishabh Jain$^1$\\
    Peter Anderson$^1$\\
    \and \\
    $^1$Georgia Institute of Technology, $^2$Macquarie University, $^3$Facebook AI Research\\
    $^4$University of Michigan, $^5$Oregon State University\\
    {\tt\small $^1$\{hagrawal9, kdexd, rishabhjain, dbatra, parikh, steflee, peter.anderson\}@gatech.edu}\\
    {\tt\small $^2$\{yufei.wang, mark.johnson\}@mq.edu.au \quad $^3$xinleic@fb.com \quad \quad \tt\normalsize \url{https://nocaps.org}}
}

\maketitle

\ificcvfinal
\renewcommand*{\thefootnote}{$\star$}
\setcounter{footnote}{1}
\footnotetext{First two authors contributed equally, listed in alphabetical order.}
\renewcommand*{\thefootnote}{\arabic{footnote}}
\setcounter{footnote}{0}
\fi

\vspace{\abstractReduceTop}
\begin{abstract}
\vspace{-0.1in}

\noindent Image captioning models have achieved impressive results on datasets containing limited visual concepts and large amounts of paired image-caption training data. However, if these models are to ever function in the wild, a much larger variety of visual concepts must be learned, ideally from less supervision. To encourage the development of image captioning models that can learn visual concepts from alternative data sources, such as object detection datasets, we present the first large-scale benchmark for this task. Dubbed `\nocaps', for novel object captioning at scale, our benchmark consists of 166,100 human-generated captions describing 15,100 images from the \openimages validation and test sets. The associated training data consists of \coco image-caption pairs, plus \openimages image-level labels and object bounding boxes. Since \openimages contains many more classes than \coco, nearly 400 object classes seen in test images have no or very few associated training captions (hence, \nocaps). We extend existing novel object captioning models to establish strong baselines for this benchmark and provide analysis to guide future work.

\end{abstract}
\vspace{\abstractReduceBot}

\vspace{\sectionReduceTop}
\section{Introduction}
\vspace{\sectionReduceTop}
\label{sec:intro}


Recent progress in image captioning, the task of generating natural language descriptions of visual content~\cite{Vinyals2015, Donahue2015, Kiros2015, Karpathy2015, Xu2015, Fang2015}, can be largely attributed to the publicly available large-scale datasets of image-caption pairs~\cite{Hodosh2013, Young2014, Chen2015MicrosoftCC} as well as steady modeling improvements~\cite{scst2016, sentinel, reviewnet, Anderson2017up-down}. However, these models generalize poorly to images in the wild~\cite{Tran2016} despite impressive benchmark performance, because they are trained on datasets which cover a tiny fraction of the long-tailed distribution of visual concepts in the real world. For example, models trained on \coco Captions~\cite{Chen2015MicrosoftCC} can typically describe images containing dogs, people and umbrellas, but not accordions or dolphins. This limits the usefulness of these models in real-world applications, such as providing assistance for people with impaired vision, or for improving natural language query-based image retrieval. 

\begin{figure}[t]
\begin{center}
\includegraphics[width=0.98\linewidth]{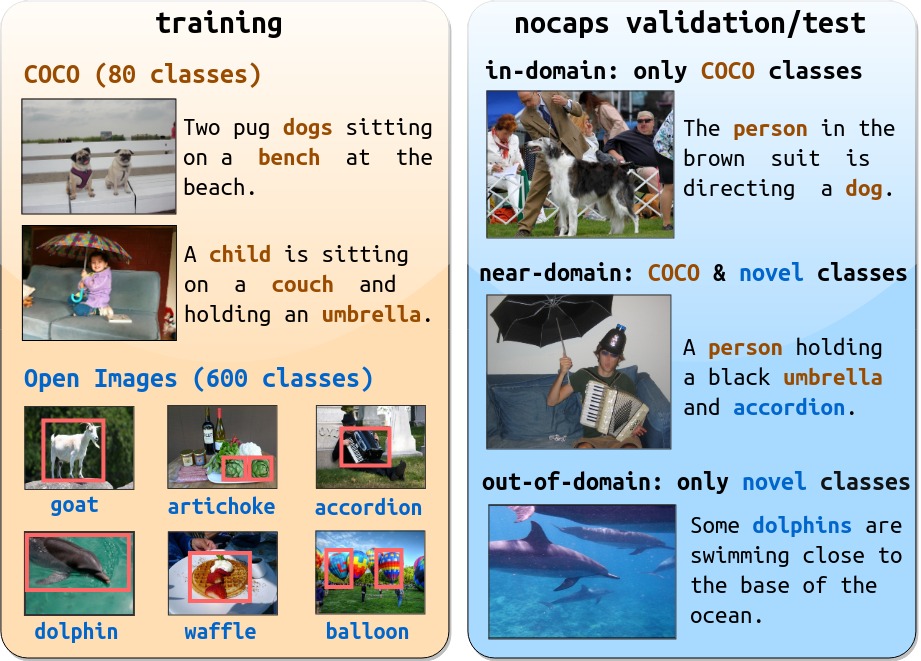}
\end{center}
\vspace{\captionReduceTop}

   \caption{The \nocaps task setup: Image captioning models must exploit the \openimages object detection dataset (bottom left) to successfully describe novel objects not covered by the \coco Captions dataset (top left). The \nocaps  benchmark (right) evaluates performance over \nocapsin, \nocapsnear and \nocapsout subsets of images containing only \coco classes, both \coco and novel classes, and only novel classes, respectively.}
\vspace{\captionReduceBot}
\label{fig:teaser}
\label{fig:long}
\label{fig:onecol}
\end{figure}

To generalize better `in the wild', we argue that captioning models should be able to leverage alternative data sources -- such as object detection datasets -- in order to describe objects not present in the caption corpora on which they are trained. Such objects which have detection annotations but are not present in caption corpora are referred to as \textit{novel objects} and the task of describing images containing novel objects is termed \textit{novel object captioning}~\cite{Hendricks2016DeepCC,Venu2016,cbs2017,yao2017incorporating,Lu2018Neural,yu2018dnoc, ps3_2018}. Until now, novel object captioning approaches have been evaluated using a proof-of-concept dataset introduced in \cite{Hendricks2016}. This dataset has restrictive assumptions -- it contains only 8 novel object classes held out from the \coco  dataset~\cite{Hendricks2016DeepCC}, deliberately selected to be highly similar to existing ones (e.g. horse is seen, zebra is novel). This has left the large-scale performance of these methods open to question. Given the emerging interest and practical necessity of this task, we introduce \nocaps, the first large-scale benchmark for novel object captioning, containing nearly 400 novel object classes.

In detail, the \nocaps benchmark consists of a validation and test set comprised of 4,500 and 10,600 images, respectively, sourced from the \openimages  object detection dataset~\cite{openimages} and annotated with 11 human-generated captions per image (10 reference captions for automatic evaluation plus a human baseline). Crucially, we provide no additional paired image-caption data for training. Instead, as illustrated in Figure~\ref{fig:onecol}, training data for the \nocaps benchmark is image-caption pairs from the \coco Captions 2017~\cite{Chen2015MicrosoftCC} training set (118K images containing 80 object classes), plus the \openimages V4 object detection training set (1.7M images annotated with bounding boxes for 600 object classes and image labels for 20K categories).

To be successful, image captioning models may utilize \coco paired image-caption data to learn to generate syntactically correct captions, while leveraging the massive \openimages detection dataset to learn many more visual concepts. Our key scientific goal is to disentangle `how to recognize an object' from `how to talk about it'. After learning the name of a novel object, a human can immediately talk about its attributes and relationships. It is therefore intellectually dissatisfying that existing models, having already internalized a huge number of caption examples, can't also be taught new objects. As with previous work, this task setting is also motivated by the observation that collecting human-annotated captions is resource intensive and scales poorly as object diversity grows, while on the other hand, large-scale object classification and detection datasets already exist~\cite{Deng2009ImageNetAL, openimages} and their collection can be massively scaled, often semi-automatically~\cite{papadopoulos2017extreme, papadopoulos2016we}.

To establish the state-of-the-art on our challenging benchmark, we evaluate two of the best performing existing approaches~\cite{Lu2018Neural, cbs2017} and report their performance based on well-established evaluation metrics -- CIDEr~\cite{Vedantam2015} and SPICE~\cite{spice2016}. To provide finer-grained analysis, we further break performance down over three subsets -- \nocapsin, \nocapsnear and \nocapsout -- corresponding to the similarity of depicted objects to \coco classes. While these models do improve over a baseline model trained only on \coco Captions, they still fall well short of human performance on this task -- indicating there is still work to be done to scale to `in-the-wild' image captioning.

\noindent In summary, we make three main contributions:
\begin{compactitem}[\hspace{1pt}-]
\item We collect \nocaps{} -- the first large-scale benchmark for novel object captioning, containing $\sim$400 novel objects.
\item We undertake a detailed investigation of the performance and limitations of two existing state-of-the-art models on this task and contrast them against human performance.
\item We make improvements and suggest simple heuristics that improve the performance of constrained beam search significantly on our benchmark.
\end{compactitem}

We believe that improvements on \nocaps will accelerate progress towards image captioning in the wild. We are hosting a public evaluation server on EvalAI~\cite{EvalAI} to benchmark progress on \nocaps. For reproducibility and to spur innovation, we have also released code to replicate our experiments at: \texttt{\textbf{https://github.com/nocaps-org}}.

\vspace{\sectionReduceTop}
\section{Related Work}
\vspace{\sectionReduceBot}
\label{sec:related}

\noindent{\textbf{Novel Object Captioning}} Novel object captioning includes aspects of both transfer learning and domain adaptation~\cite{domain-adapt}. Test images contain previously unseen, or `novel' objects that are drawn from a target distribution (in this case, \openimages~\cite{openimages}) that differs from the source/training distribution (\coco~\cite{Chen2015MicrosoftCC}). To obtain a captioning model that performs well in the target domain, 
the Deep Compositional Captioner~\cite{Hendricks2016DeepCC} and its extension, the Novel Object Captioner~\cite{Venu2016}, both attempt to transfer knowledge by leveraging object detection datasets and external text corpora by decomposing the captioning model into visual and textual components that can be trained with separate loss functions as well as jointly using the available image-caption data. 

\begin{figure*}[t]
\begin{center}
\vspace{-10pt}
\begin{minipage}[t]{.33\linewidth}
\centering
\includegraphics[width=0.90\linewidth]{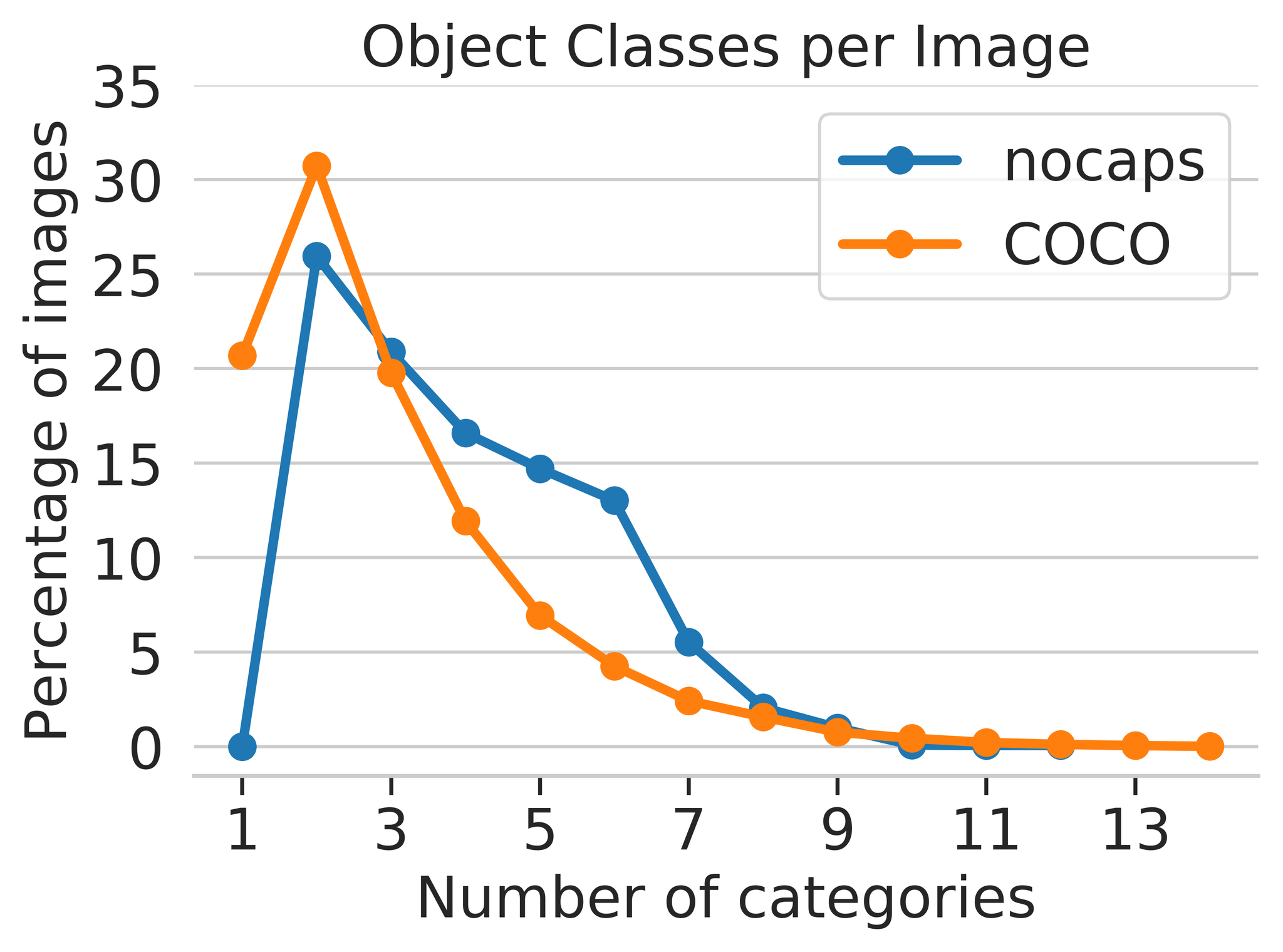}
\label{fig: nocap_vs_coco_a}
\end{minipage}%
\begin{minipage}[t]{0.33\linewidth}
\centering
\includegraphics[width=0.90\linewidth]{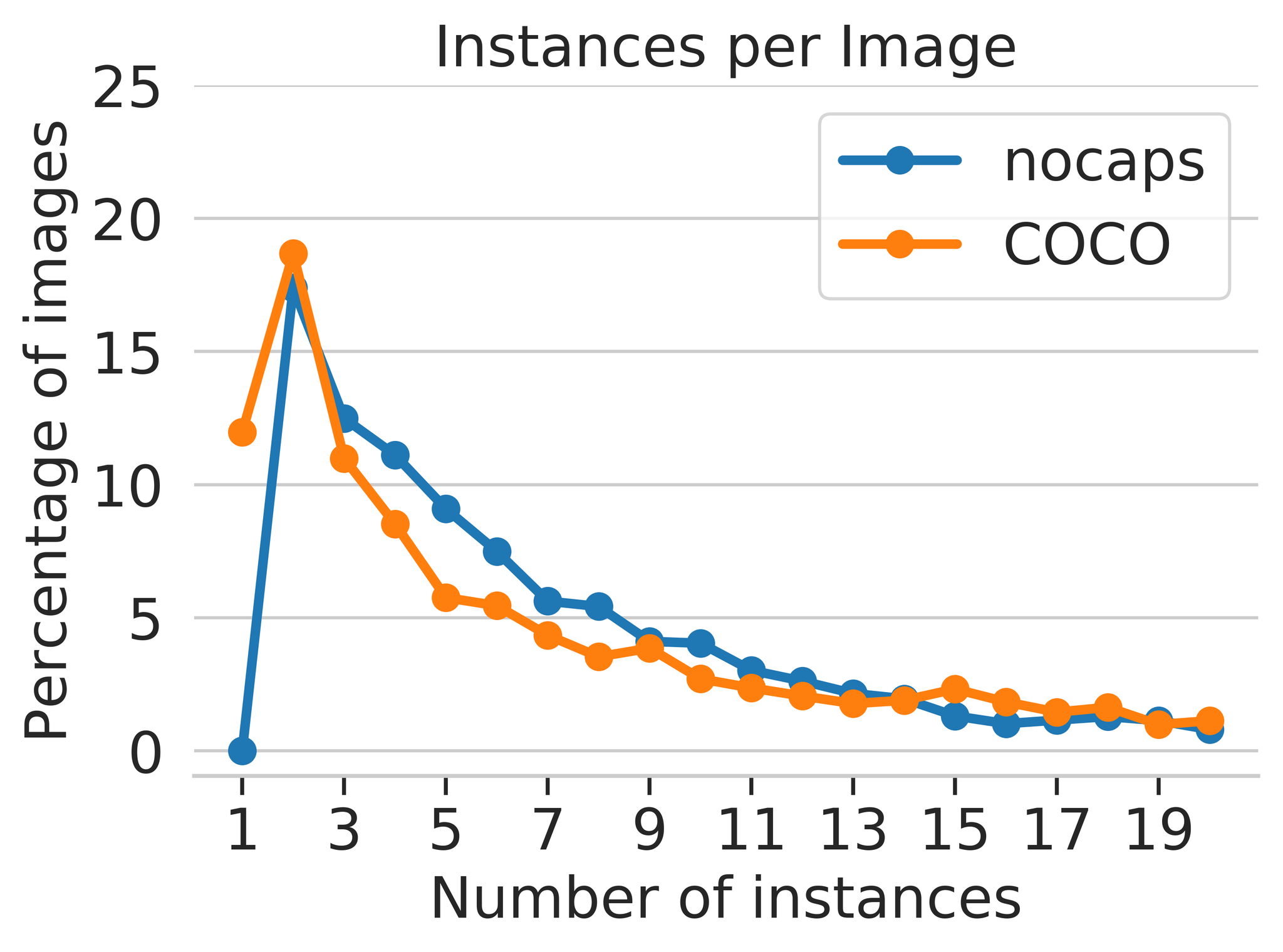}
\label{fig: nocap_vs_coco_b}
\end{minipage}
\begin{minipage}[t]{.33\linewidth}
\centering
\includegraphics[width=0.90\linewidth]{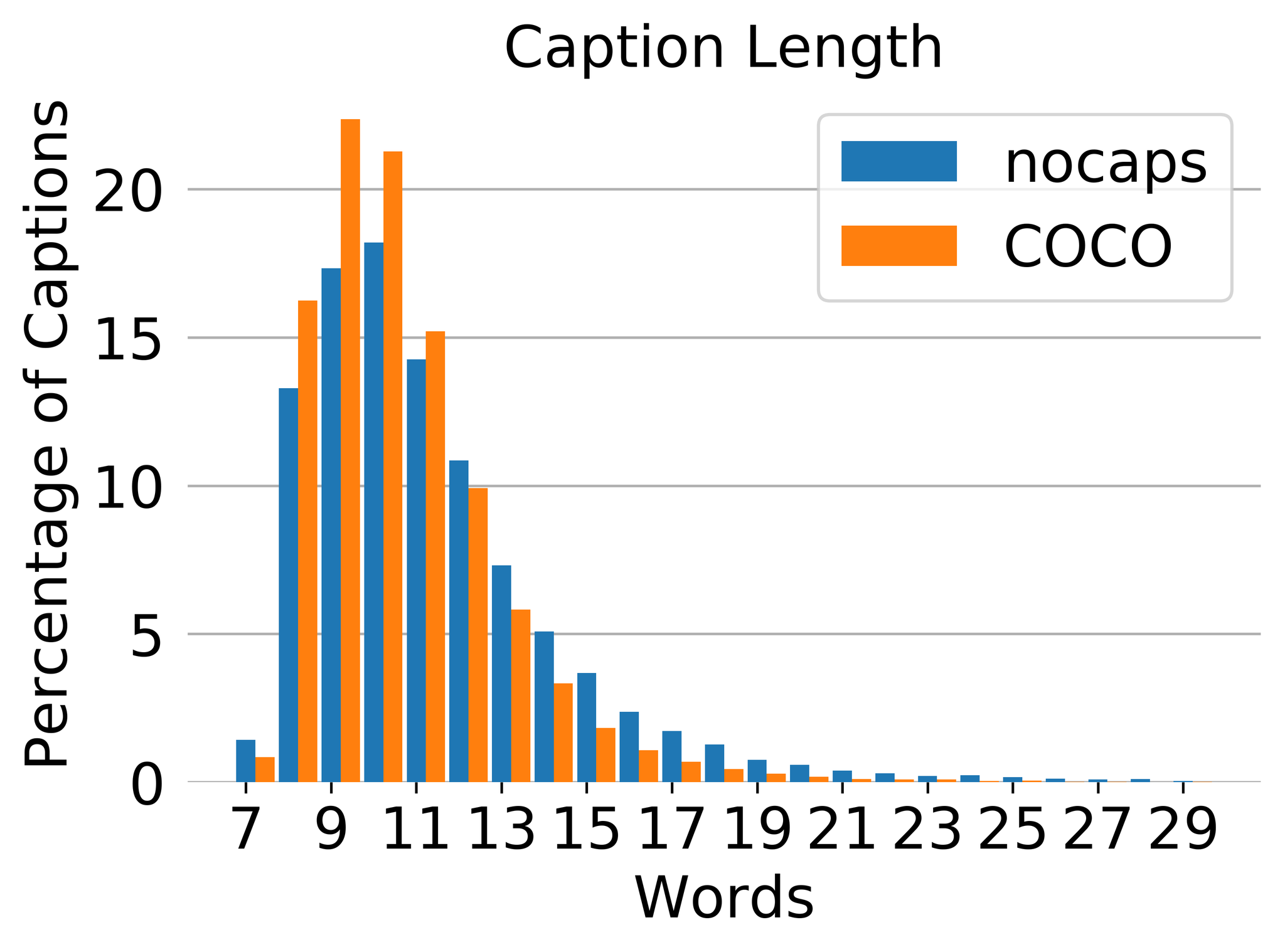}
\label{fig: nocap_vs_coco_c}
\end{minipage}
\end{center}
\vspace{\captionReduceTop}
   \caption{Compared to \coco Captions~\cite{Chen2015MicrosoftCC}, on average \nocaps images have more object classes per image (4.0 vs. 2.9), more object instances per image (8.0 vs. 7.4), and longer captions (11 words vs. 10 words). These differences reflect both the increased diversity of the underlying \openimages data~\cite{openimages}, and our image subset selection strategy (refer Section \ref{sec:data-col}).}
\label{fig:nocap_vs_coco}
\vspace{\captionReduceBot}
\end{figure*}

Several alternative approaches elect to use the output of object detectors more explicitly. Two concurrent works, Neural Baby Talk~\cite{Lu2018Neural} and the Decoupled Novel Object Captioner~\cite{yu2018dnoc}, take inspiration from Baby Talk~\cite{Kulkarni2013} and propose neural approaches to generate slotted caption templates, which are then filled using visual concepts identified by modern state-of-the-art object detectors. Related to Neural Baby Talk, the LSTM-C~\cite{yao2017incorporating} model augments a standard recurrent neural network sentence decoder with a copying mechanism which may select words corresponding to object detector predictions to appear in the output sentence. 

In contrast to these works, several approaches to novel object captioning are architecture agnostic. Constrained Beam Search ~\cite{cbs2017} is a decoding algorithm that can be used to enforce the inclusion of selected words in captions during inference, such as novel object classes predicted by an object detector. Building on this approach, partially-specified sequence supervision (PS3)~\cite{ps3_2018} uses Constrained Beam Search as a subroutine to estimate complete captions for images containing novel objects. These complete captions are then used as training targets in an iterative algorithm inspired by expectation maximization (EM)~\cite{dempster1977maximum}. 


In this work, we investigate two different approaches: Neural Baby Talk (NBT)~\cite{Lu2018Neural} and Constrained Beam Search (CBS)~\cite{cbs2017} on our challenging benchmark -- both of which recently claimed state-of-the-art on the proof-of-concept novel object captioning dataset~\cite{Hendricks2016DeepCC}.

\vspace{1mm}
\noindent{\textbf{Image Caption Datasets}} In the past, two paradigms for collecting image-caption datasets have emerged: direct annotation and filtering. Direct-annotated datasets, such as Flickr 8K~\cite{Hodosh2013}, Flickr 30K~\cite{Young2014} and \coco Captions~\cite{Chen2015MicrosoftCC} are collected using crowd workers who are given instructions to control the quality and style of the resulting captions. To improve the reliability of automatic evaluation metrics, these datasets typically contain five or more captions per image. However, even the largest of these, \coco Captions, is based only on a relatively small set of 80 object classes. In contrast, filtered datasets, such as Im2Text~\cite{Ordonez:2011:im2text}, Pinterest40M~\cite{mao2016training} and Conceptual Captions~\cite{sharma2018conceptual}, contain large numbers of image-caption pairs harvested from the web. 
These datasets contain many diverse visual concepts, but are also more likely to contain non-visual content in the description due to the automated nature of the collection pipelines. Furthermore, these datasets lack human baselines, and may not include enough captions per image for good correlation between automatic evaluation metrics and human judgments~\cite{Vedantam2015,spice2016}.

Our benchmark, \nocaps, aims to fill the gap between these datasets, by providing a high-quality benchmark with 10 reference captions per image and many more visual concepts than \coco. To the best of our knowledge, \nocaps  is the only image captioning benchmark in which humans outperform state-of-the-art models in automatic evaluation.

\vspace{-4mm}
\vspace{\sectionReduceTop}
\section{\nocaps}
\vspace{\sectionReduceBot}
\label{sec:data}

In this section, we detail the \nocaps collection process, constrast it with \coco Captions~\cite{Chen2015MicrosoftCC}, and introduce the evaluation protocol and benchmark guidelines.

\vspace{\subsectionReduceTop}
\subsection{Caption Collection}
\vspace{\subsectionReduceBot}
\label{sec:data-col}

The images in \nocaps are sourced from the \openimages V4~\cite{openimages} 
validation and test sets.~\footnote{The images used in \nocaps come from the Open Images V4 dataset and are provided under their original license (CC BY 2.0)}  \openimages is currently the largest
available human-annotated object detection 
dataset, containing 1.9M images of complex 
scenes annotated with object bounding boxes for 600 classes 
(with an average of 8.4 object instances per image in the training set).
Moreover, out of the 500 classes that are not overly broad (e.g. `clothing') or infrequent (e.g. `paper cutter'), nearly 400
are never or rarely mentioned in \coco 
Captions~\cite{Chen2015MicrosoftCC} (which we select as image-caption training data), 
making these images an ideal basis  for our benchmark.

\vspace{0.5mm}
\noindent{\textbf{Image Subset Selection}}
Since \openimages is primarily an object detection dataset, a large fraction of 
images contain well-framed iconic perspectives of single objects. Furthermore,
the distribution of object classes is highly unbalanced, with a long-tail of 
object classes that appear relatively infrequently. However, for image captioning,
images containing multiple objects and rare object co-occurrences are more
interesting and challenging. Therefore, we select subsets of images from the \openimages 
validation and test splits by applying the following sampling procedure.

First, we exclude all images for which the correct image rotation is non-zero or unknown. Next, based on the ground-truth object detection annotations, we 
exclude all images that contain only instances from a single object category. Then, to capture 
as many visually complex images as possible, we include all 
images containing more than 6 unique object classes. Finally, we iteratively select from the remaining images using a sampling procedure that encourages even representation both in terms of object classes and image complexity (based on the number of unique classes per image). Concretely, we divide the remaining images into 5 pools based on the number of unique classes present in the image (from 2--6 inclusive). Then, taking each pool in turn, we randomly sample $n$ images and among these, we select the image that when added to our benchmark results in the highest entropy over object classes. This prevents \nocaps from being overly dominated by frequently occurring object classes such as person, car or plant. In total, we select 4,500 validation images (from a total of 41,620 images in \openimages validation set) and 10,600 test images (from a total of 125,436 images in \openimages test set). On average, the selected images contain 4.0 object classes and 8.0 object instances each (see Figure \ref{fig:nocap_vs_coco}).

\begin{figure}
\centering
\begin{tabularx}{\linewidth}{XX}
	\includegraphics[width=1.0\linewidth, height=2.38cm]{figures/36daf4b907485637.jpg}&
	\includegraphics[width=1.0\linewidth, height=2.38cm]{figures/890c6c17a01f8134.jpg}\\
	\fontsize{7}{8}\selectfont \textbf{Labels}: \textcolor{NavyBlue}{Gondola}, \textcolor{OliveGreen}{Tree}, \textcolor{RedOrange}{Vehicle}&
	\fontsize{7}{8}\selectfont \textbf{Labels}: \textcolor{Maroon}{Red Panda}, \textcolor{OliveGreen}{Tree}\\
	\fontsize{8}{8}\selectfont \textbf{No Priming}: A man and a woman being transported in a boat by a sailor through canals&
	\fontsize{8}{8}\selectfont \textbf{No Priming}: A brown rodent climbing up a \textcolor{OliveGreen}{tree} in the woods.\\
	\fontsize{8}{8}\selectfont \textbf{Priming}: Some people enjoying a nice ride on a \textcolor{NavyBlue}{gondola} with a \textcolor{OliveGreen}{tree} behind them.&
	\fontsize{8}{8}\selectfont \textbf{Priming}: A \textcolor{Maroon}{red panda} is sitting in grass next to a \textcolor{OliveGreen}{tree}.\\
\end{tabularx}

    \caption{We conducted pilot studies to evaluate caption collection interfaces. Since \openimages contains rare and fine-grained classes (such as red panda, top right) we found that priming workers with the correct object categories resulted in more accurate and descriptive captions. }
	\label{fig:priming-vs-not}
\vspace{\captionReduceBot}
\vspace{\captionReduceBot}
\end{figure}

\noindent{\textbf{Collecting Image Captions from Humans}
To evaluate model-generated image captions, we collected 11 English captions for each image from a large pool of crowd-workers on Amazon Mechanical Turk (AMT). Out of 11 captions, we randomly sample one caption per image to establish human performance on \nocaps and use the remaining 10 captions as reference captions for automatic evaluation.} Prior work suggests that automatic caption evaluation metrics correlate better with human judgment when more reference captions are provided~\cite{Vedantam2015,spice2016}, motivating us to collect more reference captions than \coco (only 5 per image).

Our image caption collection interface closely resembles the interface used for collection of the \coco Captions dataset, albeit with one important difference. Since the \nocaps dataset contains more rare and fine-grained classes than \coco, in initial pilot studies we found that human annotators could not always correctly identify the objects in the image. For example, as illustrated in Figure \ref{fig:priming-vs-not}, a red panda was incorrectly described as a brown rodent. We therefore experimented with priming workers by displaying the list of ground-truth object classes present in the image. To minimize the potential for this priming to reduce the language diversity of the resulting captions, the object classes were presented as `keywords', and workers were explicitly instructed that it was not necessary to mention all the displayed keywords. To reduce clutter, we did not display object classes which are classified in \openimages as parts, e.g.~human hand, tire, door handle. Pilot studies comparing captions collected with and without priming demonstrated that primed workers produced more qualitative accurate and descriptive captions (see Figure \ref{fig:priming-vs-not}). Therefore, all \nocaps captions, including our human baselines, were collected using this priming-modified \coco collection interface. 

To help maintain the quality of the collected captions, we used only US-based workers who had completed at least 5K previous tasks on AMT with more than 95\% approval rate. We also spot-checked the captions written by each worker and blocked workers providing low-quality captions. Captions written by these workers were then discarded and replaced with captions written by high-quality workers. Overall, 727 workers participated, writing 228 captions each on average for a grand total of 166,100 captions of \nocaps. 

\begin{table}
\begin{center}\small
\setlength{\tabcolsep}{0.7em}
\begin{tabular}{lcccc}
\toprule
\textbf{Dataset} & \textbf{1-grams} & \textbf{2-grams} & \textbf{3-grams} & \textbf{4-grams} \\ 
\midrule
\coco & \multicolumn{1}{c}{6,913} & \multicolumn{1}{c}{46,664} & \multicolumn{1}{c}{92,946} & \multicolumn{1}{c}{119,582}\\
\nocaps & \multicolumn{1}{c}{8,291} & \multicolumn{1}{c}{59,714} & \multicolumn{1}{c}{116,765} & \multicolumn{1}{c}{144,577}\\ \bottomrule
\end{tabular}%
\end{center}
\vspace{\captionReduceTop}
\caption{Unique n-grams in equally-sized (4,500 images / 22,500 captions) uniformly randomly selected subset from the \coco and \nocaps validation sets. The increased visual variety in \nocaps demands a larger vocabulary compared to \coco (1-grams), but also more diverse language compositions (2-, 3- and 4-grams).}
\label{n-gram-table}
\vspace{\captionReduceBot}
\end{table}

\vspace{\subsectionReduceTop}
\subsection{Dataset Analysis}
\vspace{\subsectionReduceBot}
\label{sec:data-analysis}

In this section, we compare our \nocaps benchmark to \coco Captions~\cite{Chen2015MicrosoftCC} in terms of both image content and caption diversity. Based on ground-truth object detection annotations, \nocaps contains images spanning 600 object classes, while \coco contains only 80. Consistent with this greater visual diversity, \nocaps contains more object classes per image (4.0 vs 2.9), and slightly more object instances per image (8.0 vs 7.4) as shown in Figure \ref{fig:nocap_vs_coco}. Further, \nocaps contains no iconic images containing just one object class, whereas $20\%$ of the \coco dataset consists of such images. Similarly, less than $10\%$ of \coco images contain more than 6 object classes, while such images constitute almost $22\%$ of \nocaps. 

Although priming the workers with object classes as keywords during data collection has the potential to reduce language diversity, \nocaps captions are nonetheless more diverse than \coco. Since \nocaps images are visually more complex than \coco, on average the captions collected
to describe these images tend to be slightly longer (11 words vs. 10 words) and more diverse than the captions in the \coco dataset. As illustrated in Table \ref{n-gram-table}, 
taking uniformly random samples over the same number of images and captions in each dataset, we show that not only do \nocaps captions utilize a larger vocabulary than \coco captions  reflecting the increased number of visual concepts present. The number of
unique 2, 3 and 4-grams is also significantly higher for \nocaps -- suggesting a greater variety of unique language compositions as well. 

\begin{figure}[t]
\centering
\includegraphics[width=0.49\linewidth]{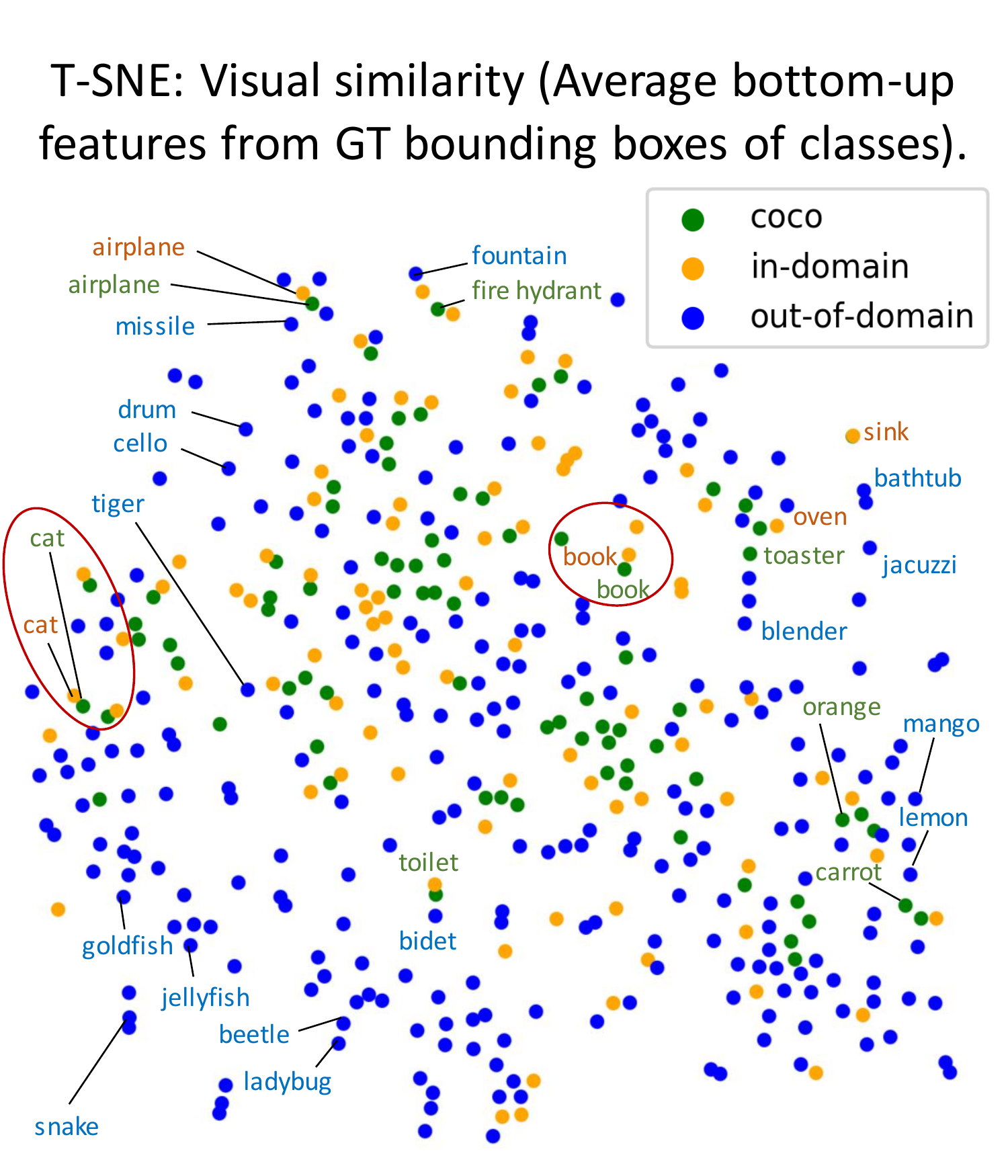}
\includegraphics[width=0.49\linewidth]{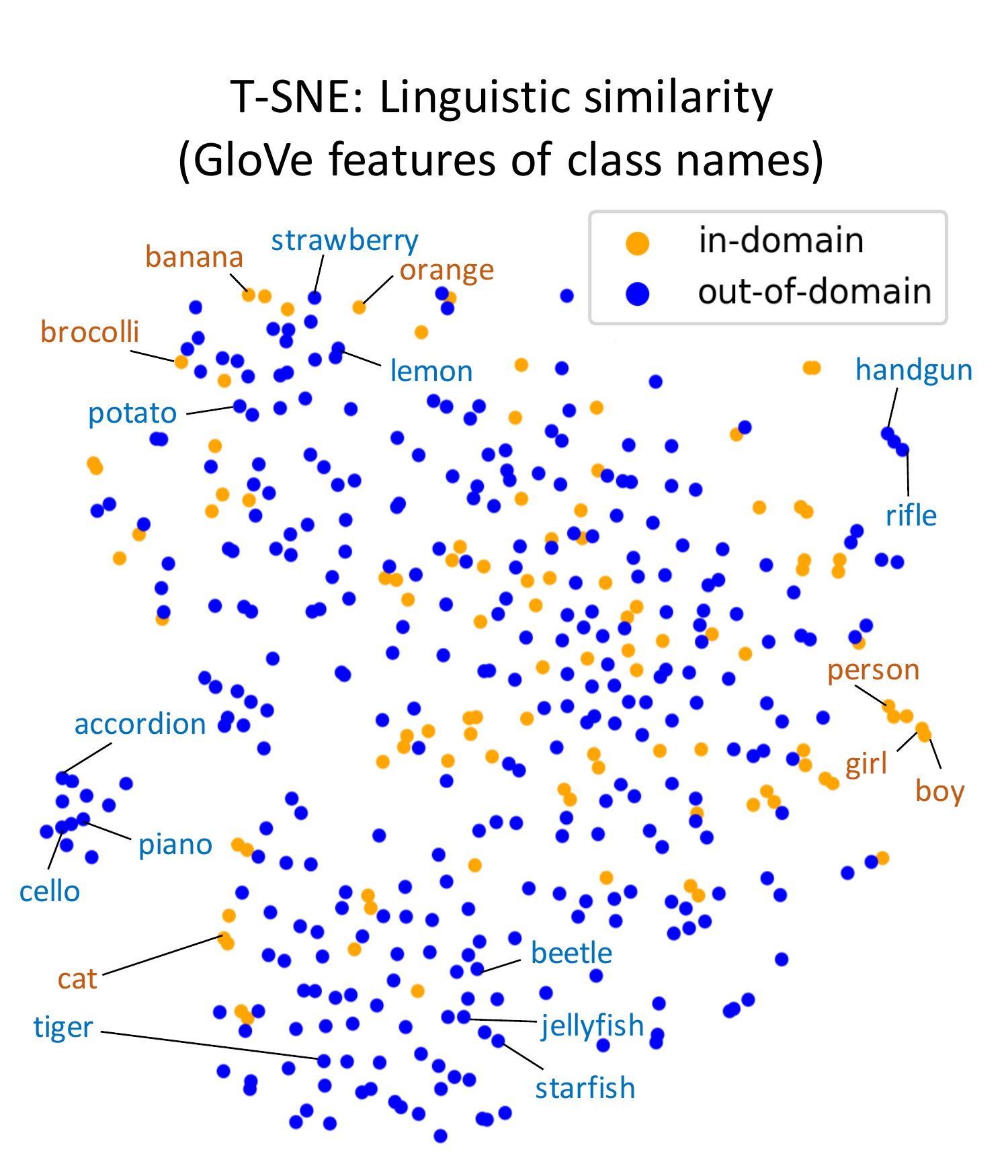}
\caption{T-SNE~\cite{maaten2008tsne} plots comparing visual (left) and linguistic (right) similarity in \coco, \nocapsin and \nocapsout classes. We observe that: (a) \nocapsin shows high visual similarity to \coco (e.g. cat, book (left)). (b) Many \nocapsout classes are visually and linguistically different from \nocapsin classes (e.g. jellyfish, beetle, cello). (c) \nocapsout also covers many visually and linguistically similar concepts to \coco, which are not well-covered in \coco (e.g. tiger, lemon).}
\label{fig:t-sne}
\vspace{\captionReduceBot}
\end{figure}

Additionally, we compare visual and linguistic similarity between \coco, \nocapsin and \nocapsout in Figure \ref{fig:t-sne}. We observe that \nocapsin classes shows high visual similarity to equivalent \coco classes (e.g. cat, book) while many \nocapsout classes are visually and linguistically different from \nocapsin classes (e.g. jellyfish, beetle, cello). \nocapsout also covers many visually and linguistically similar concepts to \coco but rarely described in \coco (e.g. tiger, lemon)

\vspace{\subsectionReduceTop}
\subsection{Evaluation}
\vspace{\subsectionReduceTop}
\label{sec:eval}

The aim of \nocaps is to benchmark progress towards models that can describe images containing visually novel concepts in the wild by leveraging other data sources. To facilitate evaluation and avoid exposing the novel object captions, we host an evaluation server for \nocaps on EvalAI~\cite{EvalAI} -- as such, we put forth these guidelines for using \nocaps:
\begin{compactitem}[\hspace{0pt}--]
\item \textbf{Do not use additional paired image-caption data collected from humans.} Improving evaluation scores by leveraging additional human-generated paired image-caption data is antithetical to this benchmark -- \textit{the only paired image-caption dataset that should be used is the \coco Captions 2017 training split}. However, external text corpora, knowledge bases, and object detection datasets
may be used during training or inference.
\item \textbf{Do not leverage ground truth object annotations.} We note that ground-truth object detection annotations are available for Open Images validation and test splits (and hence, for \nocaps). While ground-truth annotations may be used to establish performance upper bounds on the validation set, they should never be used in a submission to the evaluation server unless this is clearly disclosed.
\end{compactitem}

We anticipate that researchers may wish to investigate the limits of performance on \nocaps without any restraints on the training datasets. We therefore maintain a separate leaderboard for this purpose "\nocaps \texttt{\textbf{(XD)}}"~\footnote{XD stands for "extra data"} leaderboard.

\begin{figure}[t!]
	\centering
	\begin{center}
	\resizebox{1.0\columnwidth}{!}{
	\footnotesize
	\setlength\tabcolsep{0.5pt}
	\begin{tabularx}{\linewidth}{lXllX}
		\multicolumn{2}{c}{\normalsize{\nocapsnear}} & ~~ & \multicolumn{2}{c}{\normalsize{\nocapsout}}\\
		\multicolumn{2}{c}{\includegraphics[width=0.45\linewidth, height=2.8cm]{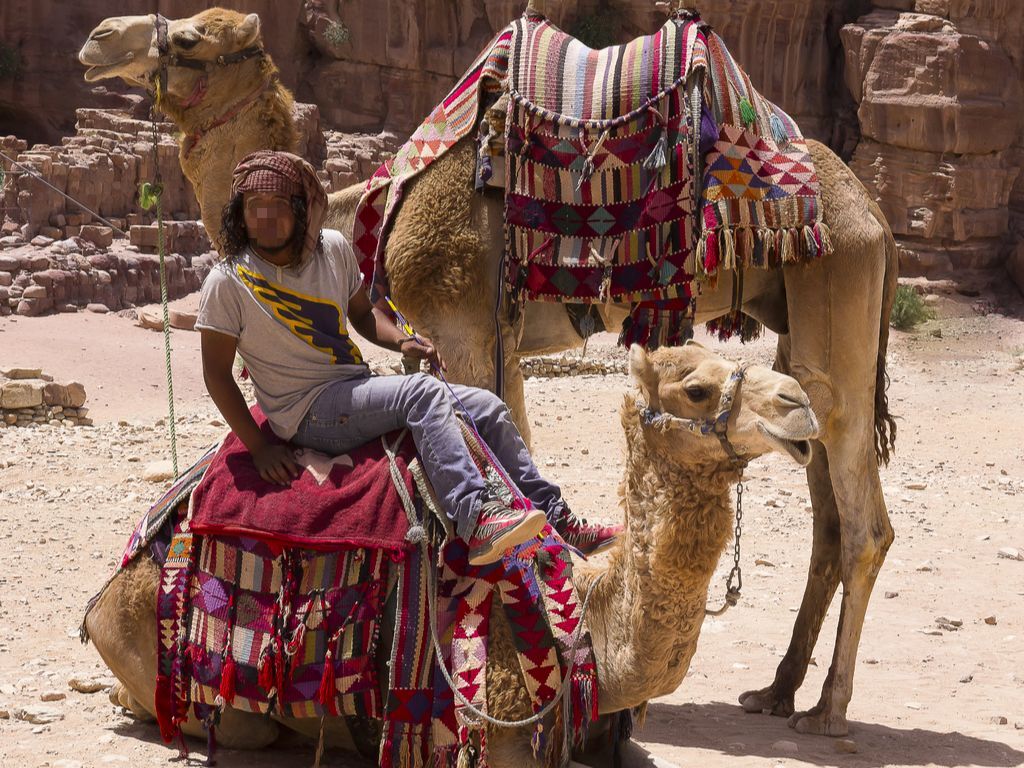}} & & \multicolumn{2}{c}{\includegraphics[width=0.45\linewidth, height=2.8cm]{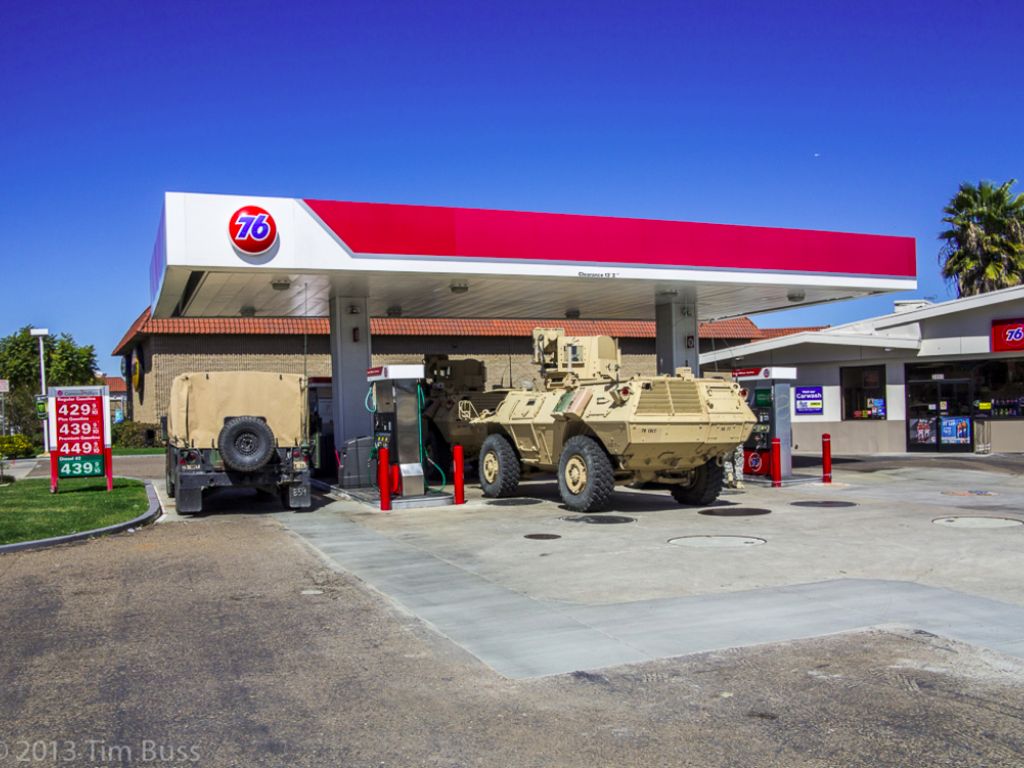}}\\
        1. & A \indomain{man} sitting in the saddle on a \outdomain{camel}. & &
        1. & A \outdomain{tank} vehicle stopped at a gas station.\\
        2. & A \indomain{person} is sitting on a \outdomain{camel} with another \outdomain{camel} behind him. & & 
		2. & A \outdomain{tank} and a military jeep at a gas station\\
		3. & A \indomain{man} with long hair and blue jeans sitting on a \outdomain{camel}.& &
		3. & A jeep and a tan colored \outdomain{tank} getting gas at a gas station. \\
		4. & \indomain{Man} sitting on a \outdomain{camel} with a standing \outdomain{camel} behind them. & &
		4. & A \outdomain{tank} and a \outdomain{truck} sit at a gas station pump.\\
		5. & Long haired \indomain{man} wearing sitting on blanket draped \outdomain{camel}& &
		5. & An Army humvee is at getting gas from the 76 gas station.\\
		6. & A \outdomain{camel} stands behind a sitting \outdomain{camel} with a \indomain{man} on its back.& &
		6. & An army \outdomain{tank} is parked at a gas station.\\
		7. & The standing \outdomain{camel} is near a sitting one with a \indomain{man} on its back.& &
		7. & A \outdomain{land vehicle} is parked in a gas station fueling.\\
		8. & Someone is sitting on a \outdomain{camel} and is in front of another \outdomain{camel}.& &
		8. & A large military vehicle at the gas pump of a gas station.\\
		9. & Two \outdomain{camels} in the dessert and a \indomain{man} sitting on the sitting one. & &
		9. & A tanker parked outside of an old gas station\\
        10. & Two \outdomain{camels} are featured in the sand with a \indomain{man} sitting on one of the seated \outdomain{camels}.& &
		10. & Multiple military vehicles getting gasoline at a civilian gas station.\\
	\end{tabularx}}
	\end{center}\vspace{-5pt}
	\caption{Examples of \nocapsnear and \nocapsout images from the \nocaps validation set. The image on the left belongs to the \nocapsnear subset (\indomain{COCO} and \outdomain{Open Images} categories), while the image on the right belongs to \nocapsout subset (only \outdomain{Open Images} categories).}
	\label{fig:examples}
	\vspace{-15pt}
\end{figure}

\noindent\textbf{Metrics}
As with existing captioning benchmarks, we rely on automatic metrics to evaluate the quality of model-generated captions. We focus primarily on CIDEr~\cite{Vedantam2015} and SPICE~\cite{spice2016}, which have been shown to have the strongest correlation with human judgments \cite{Liu_2017_ICCV} and have been used in prior novel object captioning work \cite{Hendricks2016,ps3_2018,Lu2018Neural}, but we also report Bleu~\cite{Papineni2002}, Meteor~\cite{Lavie2007} and ROUGE~\cite{Lin2004}. These metrics test whether models mention novel objects accurately \cite{Vinyals2015} as well as describe them fluently \cite{Lavie2007}.It is worth noting that the absolute scale of these metrics is not comparable across datasets due to the differing number of reference captions and corpus-wide statistics.

\noindent\textbf{Evaluation Subsets} We further break down performance on \nocaps over three subsets of the validation and test splits corresponding to varied `nearness' to \coco.

To determine these subsets, we manually map the 80 \coco classes to \openimages classes. We then select an additional 39 \openimages classes that are not \coco classes, but are nonetheless mentioned more than 1,000 times in the \coco captions training set (e.g. `table', `plate' and `tree'). We classify these 119 classes as in-domain relative to COCO. There are 87 \openimages classes that are not present in \nocaps\footnote{These classes are not included either because they are not present in the underlying \openimages val and test splits, or because they got filtered out by our image subset selection strategy favoring more complex images.}. The remaining 394 classes are out-of-domain. Image subsets are then determined as follows:
\setlist{nosep}
\begin{compactitem}[\hspace{0pt}--]
\item \nocapsin images contain only objects belonging to in-domain classes. Since these objects have been described in the paired image-caption training data, we expect caption models trained only on \coco to perform reasonably well on this subset, albeit with some negative impact due to image domain shift. This subset contains 1,311 test images (13K captions).

\item \nocapsnear images contain both in-domain and out-of-domain object classes. These images are more challenging for \coco trained models, especially when the most salient objects in the image are novel. This is the largest subset containing 7,406 test images (74K captions).

\item \nocapsout images do not contain any in-domain classes, and are visually very distinct from \coco images. We expect this subset to be the most challenging and models trained only on \coco data are likely to make `embarrassing errors'~\cite{Liu_2017_ICCV} on this subset, reflecting the current performance of \coco trained models in the wild. This subset contains 1,883 test images (19K captions).
\end{compactitem}

\vspace{\sectionReduceTop}
\section{Experiments}
\vspace{\sectionReduceBot}
\label{sec:exp}

\begin{table*}
\begin{center}
\vspace{5pt}
\footnotesize
\setlength\tabcolsep{2pt}
\renewcommand{\arraystretch}{1.2}
\begin{tabularx}{\textwidth}{lYYlYYlYYlccYYYY}
\toprule
 & \multicolumn{14}{c}{\textbf{\nocaps test}} \\ 
 \cmidrule{2-16}
 
 & \multicolumn{2}{c}{\nocapsin} && \multicolumn{2}{c}{\nocapsnear} && \multicolumn{2}{c}{\nocapsout}  && \multicolumn{6}{c}{Overall} \\
 \cmidrule{2-3} \cmidrule{5-6} \cmidrule{8-9} \cmidrule{11-16}
Method & CIDEr & SPICE && CIDEr & SPICE && CIDEr & SPICE && Bleu-1 & Bleu-4 & Meteor & ROUGE\_L & CIDEr & SPICE\\ 
\midrule
\updown      & 74.3 & 11.5 &  &  56.9 & 10.3 & & 30.1  & 8.1 &  & 74.0 & 19.2 & 23.0 & 50.9 & 54.3 & 10.1 \\
\updown + ELMo + CBS & 76.0 & 11.8 &  & 74.2 & 11.5 &  & 66.7 &  9.7 &  & 76.6  & 18.4 & 24.4 & 51.8 & 73.1 & 11.2 \\
\midrule
NBT & 60.9 & 9.8 &  & 53.2 & 9.3 &  & 48.7 & 8.2 &  & 72.3 & 14.7 & 21.5 & 48.9 & 53.4 & 9.2 \\
NBT + CBS & 63.0 & 10.1 &  & 62.0 & 9.8 &  & 58.5 & 8.8 &  & 73.4 & 12.9 & 22.1 & 48.7 & 61.5 & 9.7 \\
\midrule
Human & \textbf{80.6} & \textbf{15.0} &  & \textbf{84.6} & \textbf{14.7} &  & \textbf{91.6} & \textbf{14.2} &  & \textbf{76.6} & \textbf{19.5} & \textbf{28.2} & \textbf{52.8} & \textbf{85.3} & \textbf{14.6} \\
\bottomrule
q\end{tabularx}
\end{center}
\vspace{\captionReduceTop}
\caption{Single model image captioning performance on the \nocaps test split. We evaluate four models, including the \updown model~\cite{Anderson2017up-down} trained only on \coco, as well as three model variations based on constrained beam search (CBS)~\cite{cbs2017} and Neural Baby Talk (NBT)~\cite{Lu2018Neural} that leverage the \openimages training set.} 
\label{tab:result-test}
\vspace{0pt}
\end{table*}

\begin{table*}
\begin{center}
\footnotesize
\setlength\tabcolsep{0pt}
\renewcommand{\arraystretch}{1.2}
\begin{tabularx}{\textwidth}{c@{~~}lYYYYYlYYYYYYYY}
\toprule
& & \multicolumn{5}{c}{\textbf{\coco val 2017}} &\multicolumn{1}{c}{\textbf{~~~~}}& \multicolumn{8}{c}{\textbf{\nocaps val}} \\ 
 \cmidrule{3-7}\cmidrule{9-16}
& & \multicolumn{5}{c}{Overall} &\multicolumn{1}{c}{\textbf{}}& \multicolumn{2}{c}{\nocapsin} & \multicolumn{2}{c}{\nocapsnear} & \multicolumn{2}{c}{\nocapsout}  & \multicolumn{2}{c}{Overall} \\
& Method & Bleu-1 & Bleu-4 & Meteor & CIDEr & SPICE &\multicolumn{1}{c}{\textbf{}}& CIDEr & SPICE & CIDEr & SPICE & CIDEr & SPICE & CIDEr & SPICE \\ \midrule

\textbf{\small (1)} & \updown & \textbf{77.0} & \textbf{37.2} & \textbf{27.8} &  \textbf{116.2} & \textbf{21.0} & &  78.1 & 11.6 & 57.7 & 10.3 & 31.3 & 8.3 & 55.3 & 10.1 \\

\textbf{\small (2)} & \updown + CBS & 73.3 & 32.4 & 25.8 & 97.7 & 18.7 &  &  80.0 & 12.0 & 73.6 & 11.3 & 66.4 & 9.7 & 73.1 & 11.1  \\  
\textbf{\small (3)} & \updown + ELMo + CBS & 72.4 & 31.5 & 25.7 & 95.4 & 18.2 &  &  79.3 & 12.4 & 73.8 & 11.4 & 71.7 & 9.9 & 74.3 & 11.2  \\  

\textbf{\small (4)} & \updown + ELMo + CBS + GT & - & - & - & - & - &  & 84.2 & 12.6 & 82.1 & 11.9 & 86.7 & 10.6 & 83.3 & 11.8 \\
\midrule

\textbf{\small (5)} & NBT & 72.7 & 29.4 & 23.8 & 88.3 & 16.5 &  &  62.7 & 10.1 & 51.9 & 9.2 & 54.0 & 8.6 & 53.9 & 9.2  \\

\textbf{\small (6)} & NBT + CBS & 70.2 & 28.2 & 25.1 & 80.2 & 15.8 &  &  62.3 & 10.3 & 61.2 & 9.9 & 63.7 & 9.1 & 61.9 & 9.8  \\

\textbf{\small (7)} & NBT + CBS + GT & - & - & - & - & - &  & 68.9 & 10.7 & 68.6 & 10.3 & 76.9 & 9.8 & 70.3 & 10.3 \\

\midrule
\textbf{\small (8)} & Human & 66.3 & 21.7 & 25.2 & 85.4 & 19.8 && \textbf{84.4} & \textbf{14.3} & \textbf{85.0} & \textbf{14.3} & \textbf{95.7} & \textbf{14.0} & \textbf{87.1} & \textbf{14.2}\\ 
\bottomrule
\end{tabularx}
\end{center}

\vspace{\captionReduceTop}
\caption{Single model image captioning performance on the COCO and \nocaps validation sets. We begin with a strong baseline in the form of the \updown~\cite{Anderson2017up-down} 
trained on COCO captions. We then investigate decoding using Constrained Beam Search~\cite{cbs2017} based on object detections from the Open Images detector (+ CBS), as well as the impact of incorporating a pretrained language model (+ ELMo) and ground-truth object detections (+ GT), respectively. In panel 2, we review the performance of Neural Baby Talk (NBT)~\cite{Lu2018Neural}, illustrating similar performance trends. Even when using ground-truth object detections, all approaches lag well behind the human baseline on \nocaps. Note: Scores on \coco and \nocaps should not be directly compared (see Section \ref{sec:eval}). \coco human scores refer to the test split.}
\vspace{\captionReduceTop}
\label{tab:result-val}
\end{table*}

To provide an initial measure of the state-of-the-art on \nocaps, we extend and present results for two contemporary approaches to novel object captioning -- Neural Baby Talk (NBT)~\cite{Lu2018Neural} and Constrained Beam Search (CBS)~\cite{cbs2017} inference method which we apply both to NBT and to the popular \updown captioner~\cite{Anderson2017up-down}. We briefly recap these approaches for completeness but encourage readers to seek the original works for further details.

\noindent\textbf{Bottom-Up Top-Down Captioner (\updown) \cite{Anderson2017up-down}}
reasons over visual features extracted using object detectors trained on a large numbers of object and attribute classes and produces near state-of-the-art for single model captioning performance on \coco. For visual features, we use the publicly available Faster R-CNN~\cite{faster_rcnn} detector trained on \visualgenome by \cite{Anderson2017up-down} to establish a strong baseline trained exclusively on paired image-caption data.

\noindent\textbf{Neural Baby Talk (NBT) \cite{Lu2018Neural}} first generates a hybrid textual template with slots explicitly tied to specific image regions, and then fill these slots with words associated with visual concepts identified by an object detector. This gives NBT the capability to caption novel objects when combined with an appropriate pretrained object detector. To adapt NBT to the \nocaps setting, we incorporate the \openimages detector and train the language model using \visualgenome image features. We use fixed GloVe embeddings~\cite{pennington2014glove} in the visual feature representation for an object region for better contextualization of words corresponding to novel objects.

\noindent\textbf{Open Images Object Detection.}
Both CBS and NBT make use of object detections; we use the same pretrained Faster R-CNN model trained on \openimages for both. Specifically, we use a model\footnote{\texttt{tf\_faster\_rcnn\_inception\_resnet\_v2\_atrous\_oidv4}} from the Tensorflow model zoo~\cite{tensorflow} which achieves a detection mean average precision at 0.5 IoU (mAP@0.5) of 54\%.

\noindent\textbf{Constrained Beam Search (CBS)~\cite{cbs2017}}
CBS is an inference-time procedure that can force language models to include specific words referred to as constraints -- achieving this by casting the decoding problem as a finite state machine with transitions corresponding to constraint satisfaction. We apply CBS to both the baseline \updown model and NBT based on detected objects. Following~\cite{cbs2017}, we use a Finite State Machine (FSM) with 24 states to incorporate up to three selected objects as constraints, including two and three word phrases. After decoding, we select the highest log-probability caption that satisfies at least two constraints.

\noindent\textbf{Constraint Filtering} Although the original work~\cite{cbs2017} selected constraints from detections randomly, in preliminary experiments in the \nocaps setting we find that a simple heuristic significantly improves the performance of CBS.To generate caption constraints from object detections, we refine the raw object detection labels by removing 39 Open Images classes that are `parts' (e.g.~human eyes) or rarely mentioned (e.g.~mammal).  Specifically, we resolve overlapping detections (IoU $\geq$ 0.85) by removing the higher-order of the two objects (e.g.~, a `dog' would suppress a `mammal') based on the Open Images class hierarchy (keeping both if equal). Finally, we take the top-3 objects based on detection confidence as constraints.

\noindent\textbf{Language Embeddings} To handle novel vocabulary, CBS requires word embeddings or a language model to estimate the likelihood of word transitions. We extend the original model -- which incorporated GloVe~\cite{pennington2014glove} and dependency embeddings~\cite{levy2014dependency} -- to incorporate the recently proposed ELMo~\cite{N18-1202} model, which increased performance in our preliminary experiments. 
As captions are decoded left-to-right, we can only use the forward representation of ELMo as input encodings rather than the full bidirectional model as in~\cite{P18-2058,wang2018glue}. We also initialize the softmax layer of our caption decoder with that of ELMo and fix it during training to improve the model's generalization to unseen or rare words.


\noindent\textbf{Training and Implementation Details.} We train all models on the COCO training set and tune parameters on the \nocaps validation set. All models are trained with cross-entropy loss, i.e. we do not use RL fine-tuning to optimize for evaluation metrics~\cite{scst2016}.
\vspace{\sectionReduceTop}
\section{Results and Analysis}
\vspace{\sectionReduceBot}
\label{sec:res}

\begin{figure*}[t!]
	\hspace{-0.1in}\resizebox{1\textwidth}{!}{
	\centering
	\footnotesize
	\setlength{\tabcolsep}{0.6em}
	\begin{tabularx}{\textwidth}{lXXX}
		& \multicolumn{1}{c}{\nocapsin} & \multicolumn{1}{c}{\nocapsnear} & \multicolumn{1}{c}{\nocapsout} \\
		\multicolumn{1}{l}{\textbf{Method}} & \includegraphics[width=0.27\textwidth,height=3.0cm]{figures/val/1b150af36d87a735.jpg}&
		\includegraphics[width=0.27\textwidth,height=3.0cm]{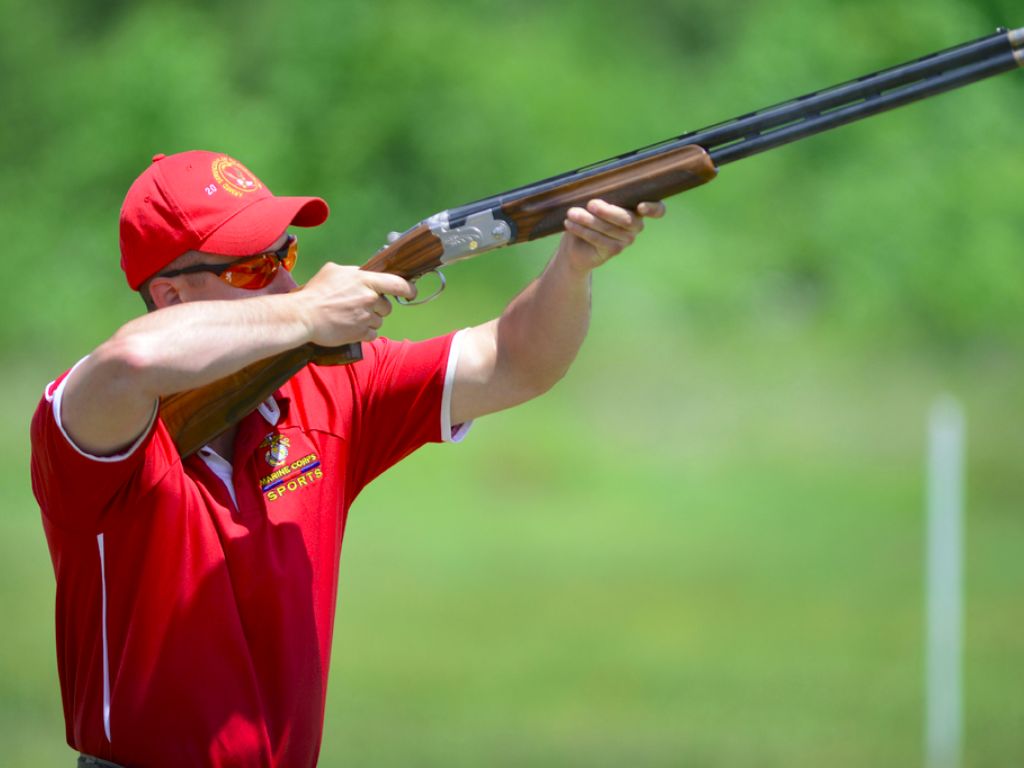} & 
		\includegraphics[width=0.27\textwidth,height=3.0cm]{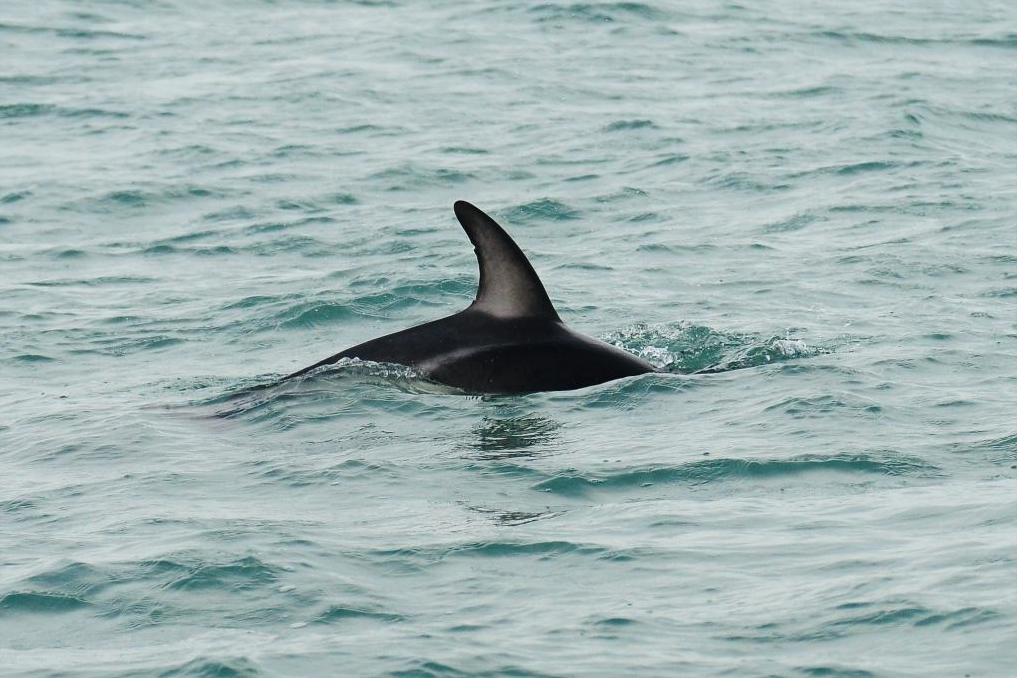}\\

		\cmidrule{1-4}
		\textbf{\updown} & A beach with chairs and umbrellas on it.&
		A man in a red shirt holding a baseball bat.&
		A bird on the ocean in the ocean.\\
		\textbf{+ ELMo} & A beach with chairs and umbrellas on it.&
		A man in a red shirt holding a baseball bat.&
		A bird that is floating on the water. \\
		\textbf{+ ELMo + CBS} & A beach with chairs and \cbs{umbrellas} and \cbs{kites}.&
		A man in a red \cbs{hat} holding a baseball \cbs{rifle}.&
		A \cbs{dolphin} swimming in the ocean on a sunny day. \\
		\textbf{\specialcell{+ ELMo + CBS\\+ GT}} & A beach with chairs and \cbs{umbrellas} on it.&
        A man in a red \cbs{hat} holding a baseball \cbs{rifle}.&
		A \cbs{whale} \cbs{dolphin} swimming in the ocean on the ocean. \\
		\midrule
		\footnotesize \textbf{NBT} & A beach with a bunch of lawn chairs and \nbt{umbrellas}.&
		A baseball \nbt{player} holding a baseball bat in the field.&
		A \nbt{dolphin} sitting in the water.\\
		\footnotesize \textbf{+ CBS} &  A beach with a bunch of \cbs{umbrellas} on a beach.&
		A baseball \nbt{player} holding a baseball \cbs{rifle} in the field.&
		A \nbt{marine mammal} sitting on a \cbs{dolphin} in the ocean. \\
		\footnotesize \textbf{+ CBS + GT} & A beach with many \cbs{umbrellas} on a beach.&
		A baseball \nbt{player} holding a baseball \cbs{rifle} in the field.&
		A black \cbs{dolphin} swimming in the ocean on a sunny day. \\
		\midrule
		\footnotesize \textbf{Human} & A couple of chairs that are sitting on a beach.&
		A man in a red hat is holding a shotgun in the air.&
		A dolphin fin is up in the water..\\
	\end{tabularx}}
	\caption{Some challenging images from \nocaps and the corresponding captions generated by our baseline models. The constraints given to the CBS are shown in \cbs{blue}, and the grounded visual words associated with NBT are shown in \nbt{red}. 
	Models perform reasonably well on \nocapsin images
	but confuse objects in \nocapsnear and \nocapsout images with visually similar \nocapsin objects, such as rifle (with baseball bat) and fin (with bird). 
	On the difficult \nocapsout images, the models generate captions with repetitions, such as "in the ocean on the ocean", and produce incoherent captions, such as "marine animal" and "dolphin" referring to the same entity in the image.}
	\label{fig:nocaps_model_predictions}
	\vspace{\captionReduceBot}
\end{figure*}

We report results on the \nocaps test set in Table \ref{tab:result-test}. While our best approach (\updown + ELMo + CBS, which is explained further below) outperforms the COCO-trained \updown baseline captioner significantly ($\sim$19 CIDEr), it still under-performs humans by a large margin ($\sim$12 CIDEr). As expected the most sizable gap occurs for \nocapsout instances ($\sim$25 CIDEr). This shows that while existing novel object captioning techniques do improve over standard models, captioning in-the-wild still presents a considerable open challenge.

In the remainder of this section, we discuss detailed results on the \nocaps and COCO validation sets (Table \ref{tab:result-val}) to help guide future work. Overall, the evidence suggests that further progress can be made through stronger object detectors and stronger language models, but open questions remain -- such as the best way to combine these elements, and the extent to which that solution should involve learning vs.~inference techniques like CBS. We align these discussions in the context of a series of specific questions below.

\begin{compactitem}[\hspace{0pt}--]
\item \noindent \textbf{Do models optimized for \nocaps maintain their performance on COCO?} We find significant gains in \nocaps performance correspond to large losses on COCO (rows \textbf{2-3 vs 1} -- dropping $\sim$20 CIDEr and $\sim$3 SPICE). Given the similarity of the collection methodology, we do not expect to see significant differences in linguistic structure between COCO and \nocaps. However, recent work has observed significant performance degradation when transferring models across datasets even when the new target dataset is an exact recreation of the old dataset~\cite{recht2018cifar}. Limiting this degradation in the captioning setting is a potential focus for future work.

\item \noindent \textbf{How important is constraint filtering?} Applying CBS greatly improves performance for both \updown and NBT (particularly on the \nocapsout captions), but success depends heavily on the quality of the constraints. Without our 39-class blacklist and overlap filtering, we find overall \nocaps validation performance falls $\sim$8 CIDEr and $\sim$3 SPICE for our \updown + ELMo + CBS model -- with most of the losses coming from the blacklisted classes. It seems likely that more sophisticated constraint selection techniques that consider image context could improve performance further.

\item \noindent \textbf{Do better language models help in CBS?} To handle novel vocabulary, CBS requires representations for the novel words. We compare using ELMo encoding (row~\textbf{3}) as described in Section \ref{sec:exp} with the setting in which word embeddings are only learned during \coco training (row~\textbf{2}). Note that in this setting the embedding for any word not found in \coco is randomly initialized. Surprisingly, the trained embeddings perform on par with the ELMo embeddings for the \nocapsin and \nocapsnear subsets, although the model with ELMo performs much better on the \nocapsout subset. It appears that even relatively rare occurrences of \nocaps object names in COCO are sufficient to learn useful linguistic models, but not visual grounding as shown by the COCO-only model's poor scores (row \textbf{1}).


\item \noindent \textbf{Do better object detectors help?} To evaluate reliance on object detections, we supply ground truth detections sorted by decreasing area to our full models (rows \textbf{4} and \textbf{7}). These ground truth detections undergo the same constraint filtering as predicted ones. Comparing to prediction-reliant models (rows \textbf{3} and \textbf{6}), we see large gains on all splits (rows \textbf{4 vs 3} -- $\sim$9 CIDEr and $\sim$0.6 SPICE gain for \updown). As detectors improve, we expect to see commensurate gains on \nocaps benchmark.

\end{compactitem}

\noindent To qualitatively assess some of the differences between the various approaches, in Figure \ref{fig:nocaps_model_predictions} we illustrate some examples of the captions generated using various model configurations. As expected, all our baseline models are able to generate accurate captions for \nocapsin images. For \nocapsnear and \nocapsout, our \updown model trained only on \coco fails to identify novel objects such as rifle and dolphin, and confuses them with known objects such as baseball bat or bird. The remaining models leverage the \openimages training data, enabling them to potentially describe these novel object classes. While they do produce more reasonable descriptions, there remains much room for improvement in both grounding and grammar. 
\vspace{\sectionReduceTop}
\section{Conclusion}
\vspace{\sectionReduceBot}
\label{sec:conclusion}

In this work, we motivate the need for a stronger and more rigorous benchmark to assess progress on the task of novel object captioning. We introduce \nocaps, a large-scale benchmark consisting of 166,100 human-generated captions describing 15,100 images containing more than 500 unique object classes and many more visual concepts. Compared to the existing proof-of-concept dataset for novel object captioning~\cite{Hendricks2016}, our benchmark contains a fifty-fold increase in the number of novel object classes that are rare or absent in training captions (394 vs 8). Further, we collected twice the number of evaluation captions per image to improve the fidelity of automatic evaluation metrics. 

We extend two recent approaches for novel object captioning to provide strong baselines for the \nocaps benchmark. While our final models improve significantly over a direct transfer from COCO, they still perform well below the human baseline -- indicating there is significant room for improvement on this task.
We provide further analysis to help guide future efforts, showing that it helps to leverage large language corpora via pretrained word embeddings and language models, that better object detectors help (and can be a source of further improvements), and that simple heuristics for determining which object detections to mention in a caption have a significant impact.

\section*{Acknowledgments}
\small{
\noindent 
We thank Jiasen Lu for helpful discussions about Neural Baby Talk. This work was supported in part by NSF, AFRL, DARPA, Siemens, Samsung, Google, Amazon, ONR YIPs and ONR Grants N00014-16-1-\{2713,2793\}. The views and conclusions contained herein are those of the authors and should not be interpreted as necessarily representing the official policies or endorsements, either expressed or implied, of the U.S. Government, or any sponsor.
}

{
\small
\bibliography{strings,bibliography}
\bibliographystyle{ieee}
}

\clearpage
\onecolumn
\setcounter{section}{0}
\vskip .1in
\begin{center}
  {\Large \bf Appendix \par}
\end{center}

\vspace{\sectionReduceTop}
In Section~\ref{sec:data_interface}, we provide the details about our data collection interface for \nocaps. Further, in Section~\ref{sec:caption_examples}, we provide some qualitative examples from \nocaps validation split. In Section~\ref{sec:additional_details}, we provide additional details in relation to the \nocaps benchmark. In Section~\ref{sec:baseline_additional}, we provide implementation details for our baseline models and finally in Section~\ref{sec:prediction_examples}, we provide examples of predicted captions on the three (\nocapsin, \nocapsnear and \nocapsout) subsets of the \nocaps validation set.

\vspace{\sectionReduceTop}
\section{Data Collection Interface} \label{sec:data_interface}
\vspace{\sectionReduceBot}
\begin{figure*}[h]
	\resizebox{1.0\textwidth}{!}{
	    \centering
	    \small
	    \includegraphics[width=1\textwidth]{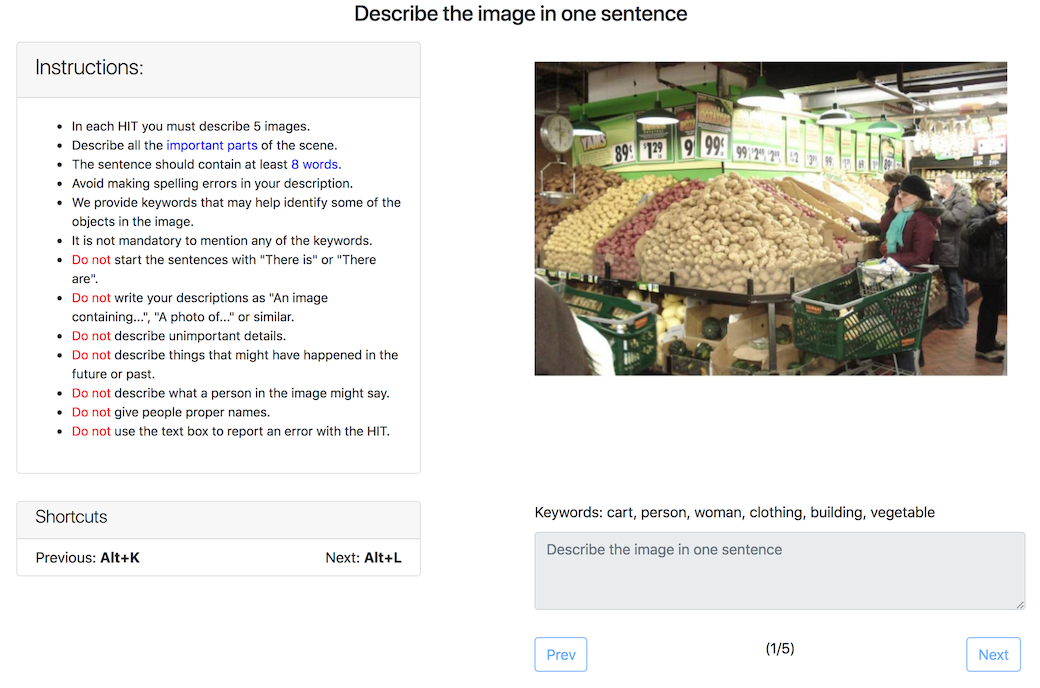}
	}
	\caption{Amazon Mechanical Turk (AMT) user interface with priming for gathering captions. The interface shows a subset of object categories present in the image as keywords. Note that the instruction explicitly states that it is not mandatory to mention any of the displayed keywords. Other instructions are similar to the interface described in \cite{Chen2015}}
	\label{fig:interface}
\end{figure*}

\clearpage

\vspace{\sectionReduceTop}
\section{Example Reference Captions from \nocaps} \label{sec:caption_examples}
\vspace{\sectionReduceBot}
\vspace{\sectionReduceTop}
\begin{figure*}[h]
	\centering
	\begin{center}
	\resizebox{0.95\textwidth}{!}{
	\footnotesize
	\setlength\tabcolsep{0.5pt}
	\begin{tabularx}{\linewidth}{lXllXllX}
		\multicolumn{2}{c}{\normalsize{\nocapsin}} & ~~ & \multicolumn{2}{c}{\normalsize{\nocapsnear}} & ~~ & \multicolumn{2}{c}{\normalsize{\nocapsout}}\\
		
    \multicolumn{2}{c}{\includegraphics[width=0.33\linewidth,height=4cm, clip]{figures/val/6e9dcdbee43b0346.jpg}}
            & 
            & 
            \multicolumn{2}{c}{\includegraphics[width=0.33\linewidth,height=4cm, clip]{figures/val/fb71b085f783c008.jpg}}
            & 
            & 
            \multicolumn{2}{c}{\includegraphics[width=0.33\linewidth,height=4cm]{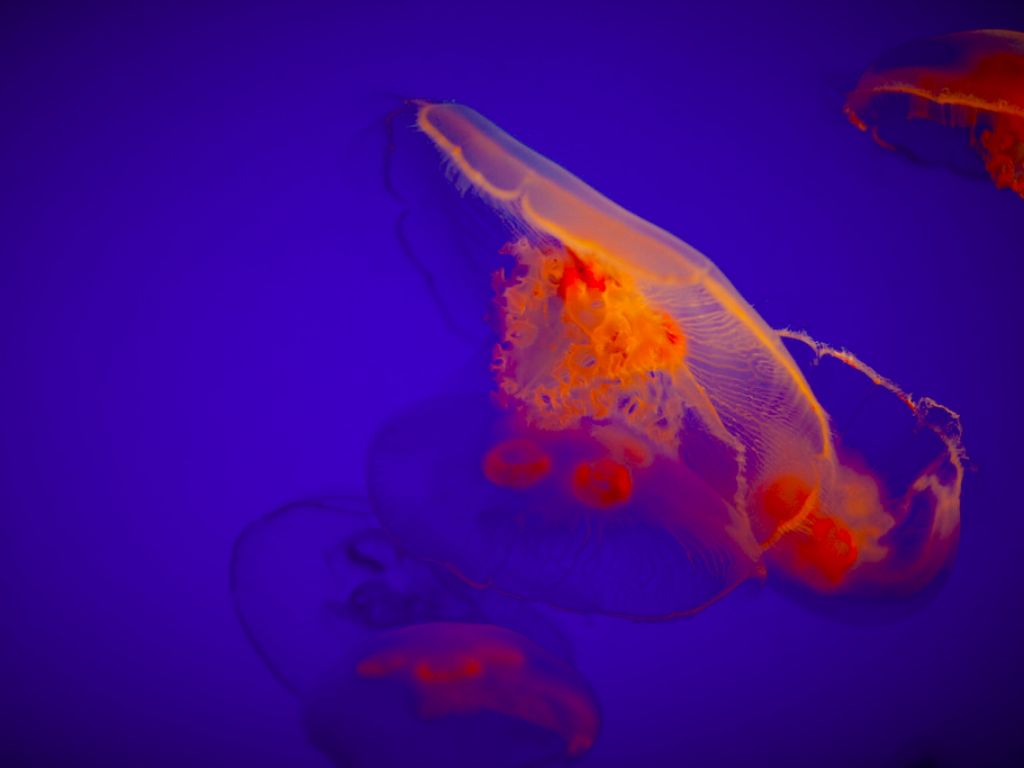}}\\

        1. & Two hardcover \indomain{\indomain{books}} are on the \indomain{table}  & &
            1. & \indomain{Men} in military uniforms playing instruments in an orchestra. & &
            1. & Some red invertebrate \outdomain{jellyfishes} in dark blue water. \\

        2. & Two magazines are sitting on a \indomain{coffee table}.  & &
            2. & Military officers play brass horns next to each other. & &
            2. & orange and clear \outdomain{jellyfish} in dark blue water\\

        3. & Two \indomain{\indomain{books}} and many crafting supplies are on this \indomain{table}. & &
            3. & Two \indomain{men} in camouflage clothing playing the \outdomain{trumpet}.  & &
            3. & A red \outdomain{jellyfish} is swimming around with other red \outdomain{jellyfish}.\\

        4. & a recipe \indomain{book} and sewing \indomain{book} on a craft \indomain{table} & &
            4. & Two \indomain{men} dressed in military outfits play the french horn. & &
            4. & Orange \outdomain{jellyfish} swimming through the water. \\

        5. & Two hardcover \indomain{\indomain{books}} are laying on a \indomain{table}. & &
            5. & Two \indomain{people} dressed in camoflauge uniforms playing musical instruments. & &
            5. & Bright orange and clear \outdomain{jellyfish} swim in open water.\\

        6. & A \indomain{table} with two different \indomain{\indomain{books}} on it. & &
            6. & A couple \indomain{people} in uniforms holding tubas by their mouths. & &
            6. & The fish is going through the very blue water.\\

        7. & Two different \indomain{books} on sewing and cooking/baking on a \indomain{table}. & &
            7. & Two \indomain{people} in uniform are playing the tuba. & &
            7. & A bright orange \outdomain{jellyfish} floating in the water.\\

        8. & Two magazine \indomain{books} are sitting on a \indomain{table} with arts and craft materials. & &
            8. & A couple of military \indomain{men} playing the french horn.  & &
            8. & Several red \outdomain{jellyfish} swimming in bright blue water.\\

        9. & A couple of \indomain{books} are on a \indomain{table}. & &
            9. & A \indomain{man} in uniform plays a French horn. & &
            9. & An orange \outdomain{jellyfish} swimming with a blue background\\

        10. & The \indomain{person} is there looking into the \indomain{book}. & &
            10. & Two \indomain{men} are playing the \outdomain{trumpet} standing nearby. & &
            10. & A very vibrantly red \outdomain{jellyfish} is seen swimming in the water.\\
            \addlinespace[0.1cm]
        \multicolumn{2}{c}{\includegraphics[width=0.33\linewidth,height=4cm]{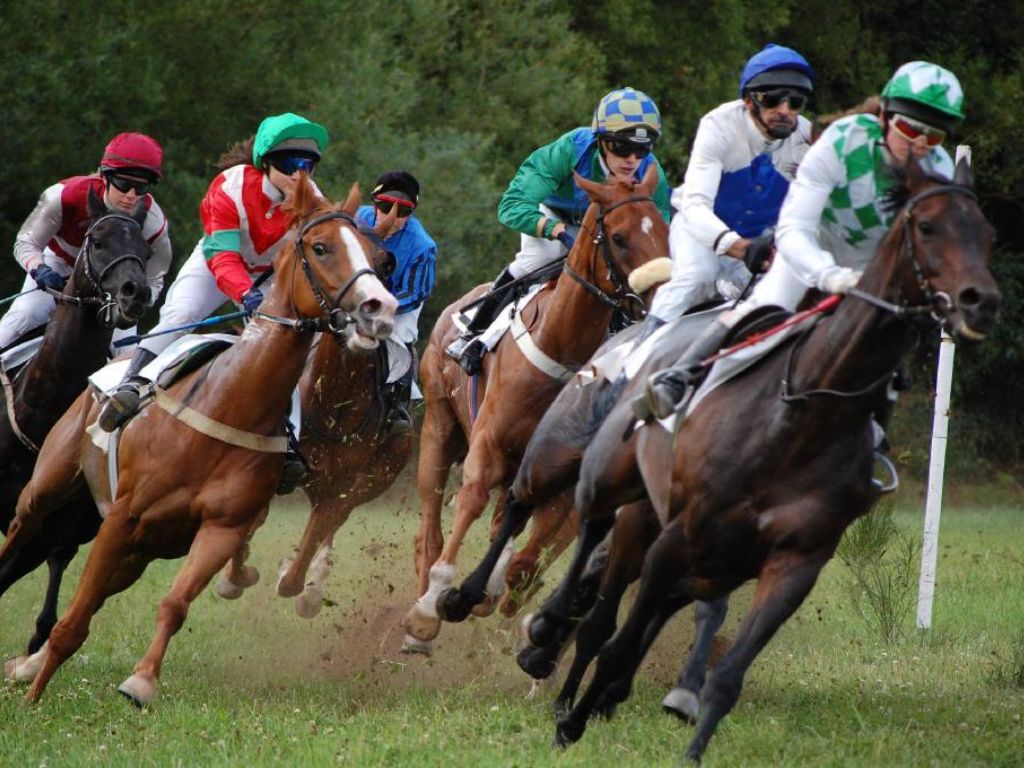}} 
            & 
            & 
            \multicolumn{2}{c}{\includegraphics[width=0.33\linewidth,height=4cm, clip]{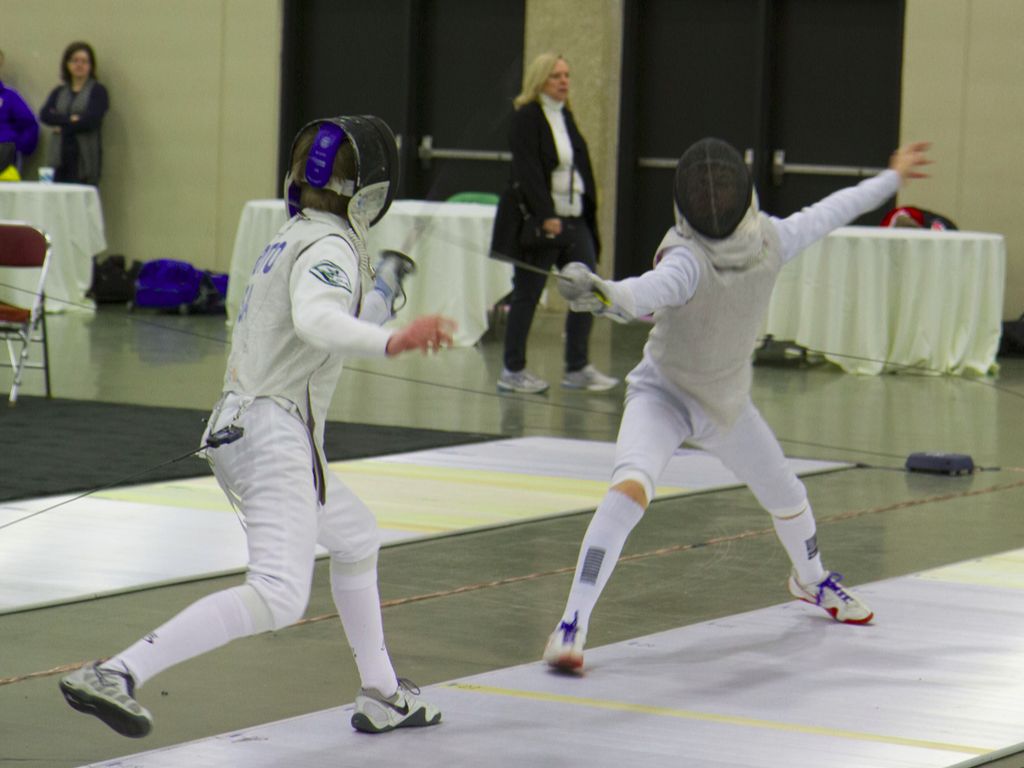}}
            & 
            & 
            \multicolumn{2}{c}{\includegraphics[width=0.33\linewidth,height=4cm, clip]{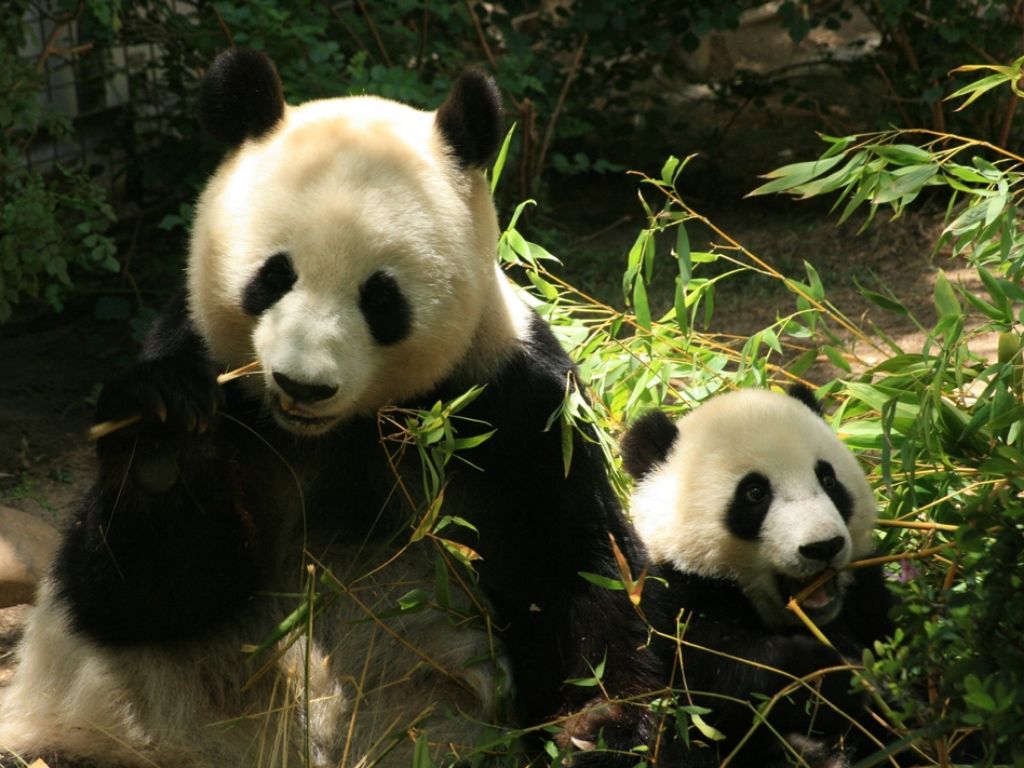}}\\

        1. & Jockeys on \indomain{horses} racing around a track.  & &
            1. & Two \indomain{people} in a fencing match with a \indomain{woman} walking by in the background. & &
            1. & A \outdomain{panda} bear sitting beside a smaller \outdomain{panda} bear.\\

        2. & Several \indomain{horses} are running in a race thru the grass. & &
            2. & Two \indomain{people} in masks fencing with each other. & &
            2. & The \outdomain{panda} is large and standing over the plant.\\
        3. & several \indomain{people} racing \indomain{horses} around a turn outside & &
            3. & Two \indomain{people} in white garbs are fencing while \indomain{people} watch. & &
            3. & Two \outdomain{panda} are eating sticks from plants. \\
        4. & Uniformed jockeys on \indomain{horses} racing through a grass field. & &
            4. & Two \indomain{people} in full gear fencing on white mat. & &
            4. & Two \outdomain{panda} bears sitting with greenery surrounding them.\\
        5. & Several \indomain{horse} jockies are riding \indomain{horses} around a turn. & &
            5. & A couple of \indomain{people} in white outfits are fencing. & &
            5. & two \outdomain{panda} bears in the bushes eating bamboo sticks \\
        6. & Six men and six \indomain{horses} are racing outside & &
            6. & Two fencers in white outfits are dueling indoors. & &
            6. & two \outdomain{pandas} sitting in the grass eating some plants\\
        7. & A group of men wearing sunglasses and racing on a horse  & &
            7. & A couple of \indomain{people} doing a fencing competition inside.  & &
            7. & two \outdomain{pandas} are eating a green leaf from a plant\\
        8. & Six \indomain{horses} with riders are racing, leaning over at an incredible angle. & &
            8. & Two \indomain{people} in white clothes fencing each other. & &
            8. & Two \outdomain{pandas} are eating bamboo in a wooded area.\\

        9. & Seveal \indomain{people} wearing goggles and helmets racing \indomain{horses}. & &
            9. & Two \indomain{people} in an room competing in a fencing competition. & &
            9. & \outdomain{Pandas} enjoy the outside and especially with a friend.\\

        10. & a row of \indomain{horses} and jockeys running in the same direction in a line & &
            10. & Two \indomain{people} in all white holding swords and fencing. & &
            10. & Two black and white \outdomain{panda} bears eating leaf stems \\
	\end{tabularx}}
	\end{center}\vspace{-10pt}
	\caption{Examples of images belonging to the \nocapsin, \nocapsnear and \nocapsout subsets of the \nocaps validation set. Each image is annotated with 10 reference captions, capturing more of the salient content of the image and improving the accuracy of automatic evaluations~\cite{Vedantam2015,spice2016}. Categories in \indomain{orange} are \nocapsin object classes while categories in \outdomain{blue} are \nocapsout classes. Note that not all captions mention the ground-truth object classes consistent with the instructions provided on the data collection interface.}
	\label{fig:examples1}
	\vspace{-15pt}
\end{figure*}

\begin{figure*}[t!]
	\centering
	\begin{center}
	\resizebox{0.95\textwidth}{!}{
	\footnotesize
	\setlength\tabcolsep{0.5pt}
	\begin{tabularx}{\linewidth}{lXllXllX}
		\multicolumn{2}{c}{\normalsize{\nocapsin}} & ~~ & \multicolumn{2}{c}{\normalsize{\nocapsnear}} & ~~ & \multicolumn{2}{c}{\normalsize{\nocapsout}}\\
		
            \multicolumn{2}{c}{\includegraphics[width=0.33\linewidth,height=4cm, clip]{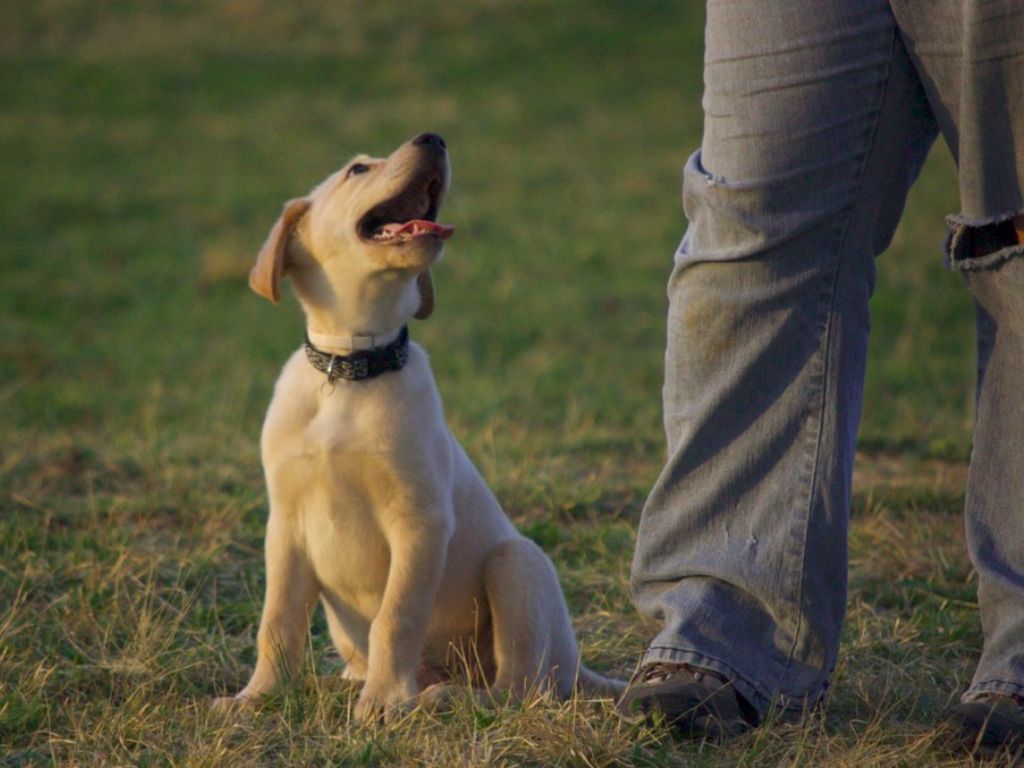}} 
            & 
            & 
            \multicolumn{2}{c}{\includegraphics[width=0.33\linewidth,height=4cm, clip]{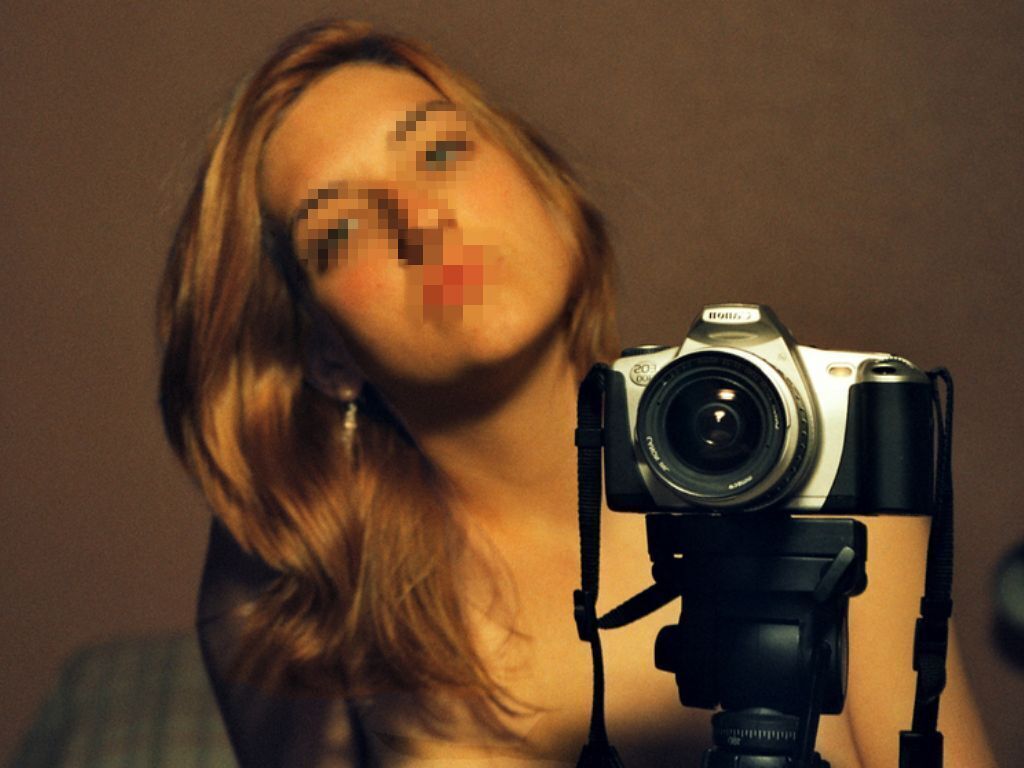}}
            & 
            & 
            \multicolumn{2}{c}{\includegraphics[width=0.33\linewidth,height=4cm, clip]{figures/val/2feba53355b4e765.jpg}}\\

        1. & A \indomain{dog} sitting beside a \indomain{man} walking on the lawn  & &
            1. & A \indomain{woman} is sitting  with a \outdomain{camera} in front of her & &
            1. & Some decorations are have red lights you can see at night.\\

        2. & A small \indomain{dog} looking up at a \indomain{person} standing next to him. & &
            2. & A beautiful brown haired \indomain{woman} next to a \outdomain{camera}. & &
            2. & A couple of red \outdomain{lanterns} floating in the air.\\

        3. & A young \indomain{dog} looks up at their owner. & &
            3. & A \indomain{woman} poses to take a picture in a mirror. & &
            3. & Many red Chinese \outdomain{lanterns} are hung outside at night\\

        4. & A little puppy looking up at a \indomain{person}.  & &
            4. & A \indomain{women} with brown hair holding a \outdomain{camera}. & &
            4. & Floating lighted \outdomain{lanterns} on a dark night in the city. \\

        5. & A \indomain{dog} sitting on the grass next to a human. & &
            5. & A \indomain{woman} behind a \outdomain{camera} on a \outdomain{tripod}. & &
            5. & Dozens of glowing paper \outdomain{lanterns} floating off into the sky.\\

        6. & A \indomain{dog} is looking up at the \indomain{person} who is wearing jeans.   & &
            6. & A \indomain{woman} sits with her head leaned behind a \outdomain{camera}. & &
            6. & A black night sky with red, bright floating \outdomain{lanterns}.\\

        7. & The tan \indomain{dog} sits patiently beside the \indomain{person}.l & &
            7. & A girl looking pity behind a \outdomain{camera} on a \outdomain{tripod}. & &
            7. & Red \outdomain{lanterns} floating up to the dark night sky.\\

        8. & The white \indomain{dog} is sitting in the grass by a \indomain{person} who is standing up.   & &
            8. & A \indomain{woman} sits and tilts her head while behind a \outdomain{camera}. & &
            8. & Chinese \outdomain{lanterns} that are red are floating into the sky.\\

        9. & The tan \indomain{dog} happily accompanies the human on the grass. & &
            9. & A \indomain{woman} is sitting behind a \outdomain{camera} with \outdomain{tripod}.  & &
            9. & This town has many lit Chinese \outdomain{lanterns} hanging between the buildings.\\

        10. & A \indomain{dog} is on the grass is looking to a \indomain{person} & &
            10. & A \indomain{woman} with a \outdomain{camera} in front of her. & &
            10. & The street is filled with light from hanging \outdomain{lanterns}.\\
        \addlinespace[0.2cm]
        \multicolumn{2}{c}{\includegraphics[width=0.33\linewidth,height=4cm, clip]{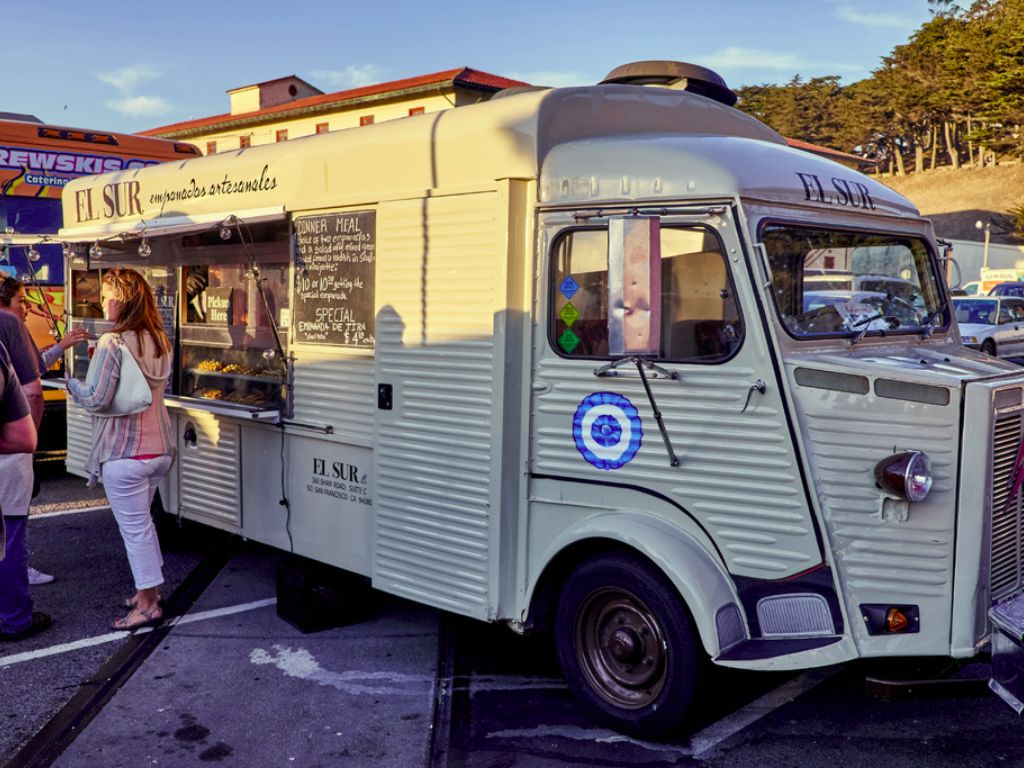}}
            & 
            & 
            \multicolumn{2}{c}{\includegraphics[width=0.33\linewidth,height=4cm, clip]{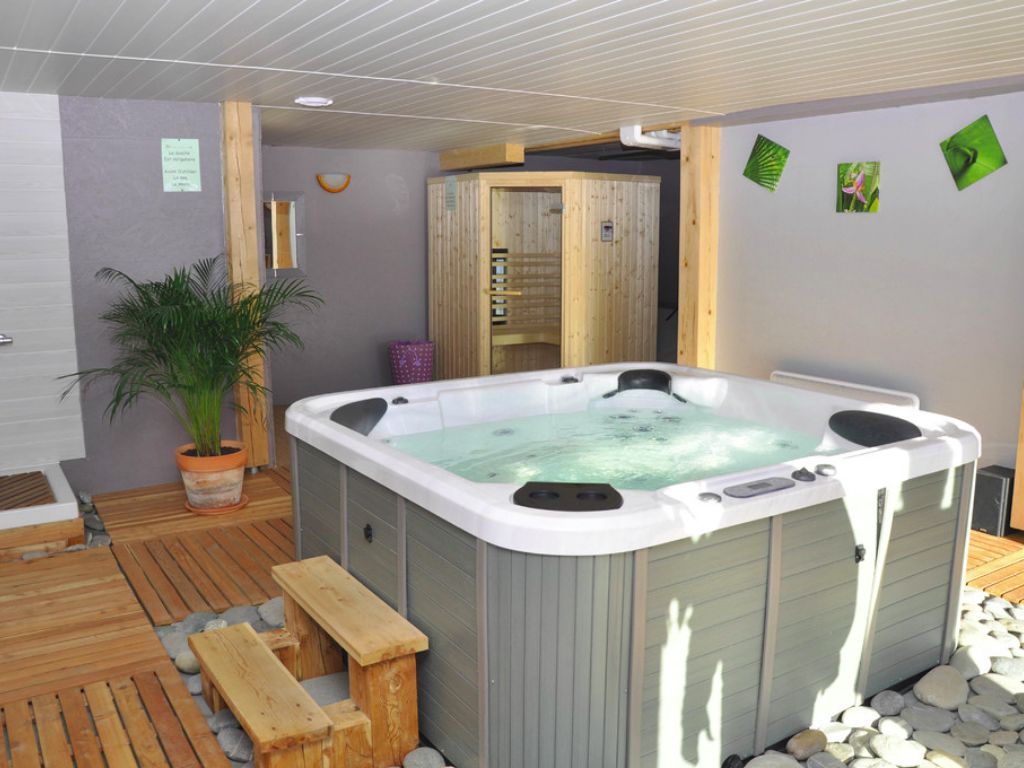}}
            & 
            & 
            \multicolumn{2}{c}{\includegraphics[width=0.33\linewidth,height=4cm, clip]{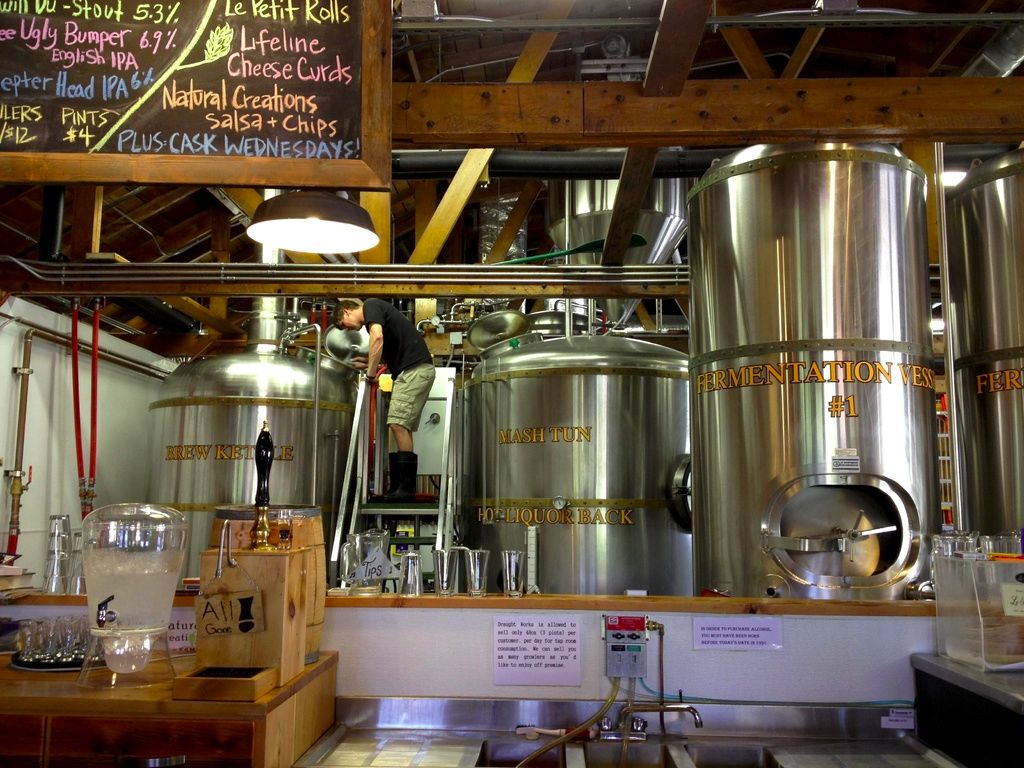}}\\

        1. & \indomain{people} are standing on the side of a food \indomain{truck}  & &
            1. & A room with a hot tub and sauna.  & &
            1. & Large silver \outdomain{tanks} behind the counter at a restaurant. \\

        2. & A food \indomain{truck} parked with \indomain{people} standing in line.  & &
            2. & A white hot tub is next to some wood. & &
            2. & Shiny metal \outdomain{containers} with writing are beside each other.\\

        3. & \indomain{people} standing outside and ordering from a food \indomain{truck} in the daytime. & &
            3. & A \outdomain{jacuzzi} sitting on rocks inside of a patio. & &
            3. & A brewery with big, silver, metal \outdomain{containers} and a sign.\\

        4. & The food \indomain{truck} has a line of \indomain{people} in front of the window. & &
            4. & A hot tub sits in the middle of the room. & &
            4. & A brew station inside of a restaurant. \\

        5. & A \indomain{woman} standing in front of a food \indomain{truck}. & &
            5. & A \outdomain{jacuzzi} sitting near some rocks and a sauna & &
            5. & Large steel breweries sit behind a chalkboard displaying different food and drink deals.\\

        6. & A food  \indomain{truck} outside of a small business  with several \indomain{people} eating  & &
            6. & A hot tub in a room with wooden flooring. & &
            6. & A cabinetry with big tin cans and a chalkboard on the top\\

        7. & \indomain{people} stand in line to get food from a food \indomain{truck}. & &
            7. & A room is shown with a hot tub, decorative plants and some paintings ont he wall. & &
            7. & A man works on machinery inside a brewery.\\

        8. & A large metal \indomain{truck} serving food to \indomain{people} in a parking lot. & &
            8. & A room with a large hot tub and a sauna. & &
            8. & The many silver \outdomain{tanks} are used for beverage making.\\

        9. & \indomain{men} and \indomain{women} speaking in front of a grey food \indomain{truck} that is open for business. & &
            9. & A water filled \outdomain{jacuzzi} surrounded by smooth river rocks and a wooden deck. & &
            9. & A menu is hanging above a craft brewery.\\

        10. & \indomain{woman} wearing jeans in front of the \indomain{truck} & &
            10. & A white and grey \outdomain{jacuzzi} around rock building  & &
            10. & A man peers at a brewing tank while standing on a step ladder. \\
	\end{tabularx}}
	\end{center}\vspace{-10pt}
	\caption{More examples of images belonging to the \nocapsin, \nocapsnear and \nocapsout subsets of the \nocaps validation set. Each image is annotated with 10 reference captions, capturing more of the salient content of the image and improving the accuracy of automatic evaluations~\cite{Vedantam2015,spice2016}. Categories in \indomain{orange} are \nocapsin object classes while categories in \outdomain{blue} are \nocapsout classes. Note that not all captions mention the ground-truth object classes consistent with the instructions provided on the data collection interface.}
	\label{fig:examples2}
	\vspace{-15pt}
\end{figure*}

\clearpage

\vspace{\sectionReduceTop}
\section{Additional Details about \nocaps Benchmark} \label{sec:additional_details}
\vspace{\sectionReduceBot}
\subsection{Evaluation Subsets}

\begin{multicols*}{2}

As outlined in Section 3.3 of the main paper, to determine the \nocapsin, \nocapsnear and \nocapsout subsets of \nocaps, we first classify Open Images classes as either \nocapsin or \nocapsout with respect to COCO. To identify the \nocapsin Open Images classes, we manually map the 80 COCO classes to Open Images classes. We then select an additional 39 Open Images classes that are not COCO classes, but are nonetheless mentioned more than 1,000 times in the \coco captions training set (e.g. `table', `plate' and `tree'), and we classify all 119 of these classes as \nocapsin. The remaining classes are considered to be \nocapsout.

To put this in perspective, in Figure \ref{fig:out_domain_hist} we plot the number of mentions of both the \nocapsin classes (in orange) and the \nocapsout classes (in blue) in the COCO Captions training set using a log scale. As intended, the \nocapsin object classes occur much more frequently in \coco Captions compared to \nocapsout object classes. However, it is worth noting that the \nocapsout are not necessarily absent from \coco Captions, but they are relatively infrequent which makes these concepts hard to learn from \coco.

\paragraph{Open Images classes ignored during image subset selection:} We also note that 87 Open Images classes were not considered during the image subset selection procedure to create \nocaps, for one of the following reasons:

\begin{itemize}
    \item \textbf{Parts:} In our image subset selection strategy (refer Section 3.1 of the main paper), we ignored `part' categories such as `vehicle registration plate', `wheel', `human-eye', which always occur with parent categories such car, person;
    \item \textbf{Super-categories:} Our image subset selection strategy also ignored super-categories such as `sports equipment', `home appliance', `auto part' which are often too broad and subsumes both \coco and \openimages categories;
    \item \textbf{Solo categories:} Certain categories such as `chime' and `stapler' did not appear in images alongside any other classes, and so were filtered out by our image subset selection strategy; and
    \item \textbf{Rare categories:} Some rare categories such as `armadillo', `pencil sharpener' and `pizza cutter' do not actually occur in the underlying \openimages val and test splits.
\end{itemize}
\hfill \break

\begin{Figure}
	    \centering
	    \small
\includegraphics[width=0.82\textwidth]{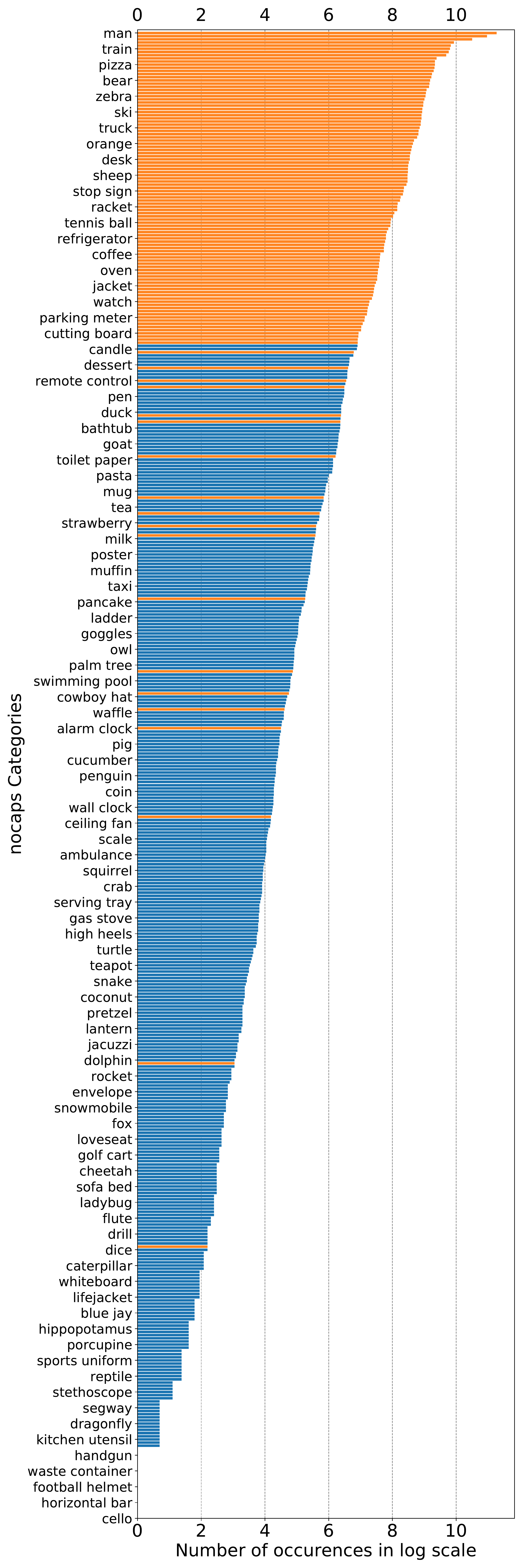}
	\captionof{figure}{Histogram of mentions in the \coco Captions training set for various Open Images object classes. In \nocaps, classes in orange are considered to be \nocapsin while classes in blue are classified as \nocapsout. \textit{Zoom in for details.} }
	\label{fig:out_domain_hist}
\end{Figure}

\end{multicols*}
\clearpage

\subsection{T-SNE Visualization in Visual Feature Embedding Space}
\begin{figure}[h]
\begin{center}
\includegraphics[width=\linewidth]{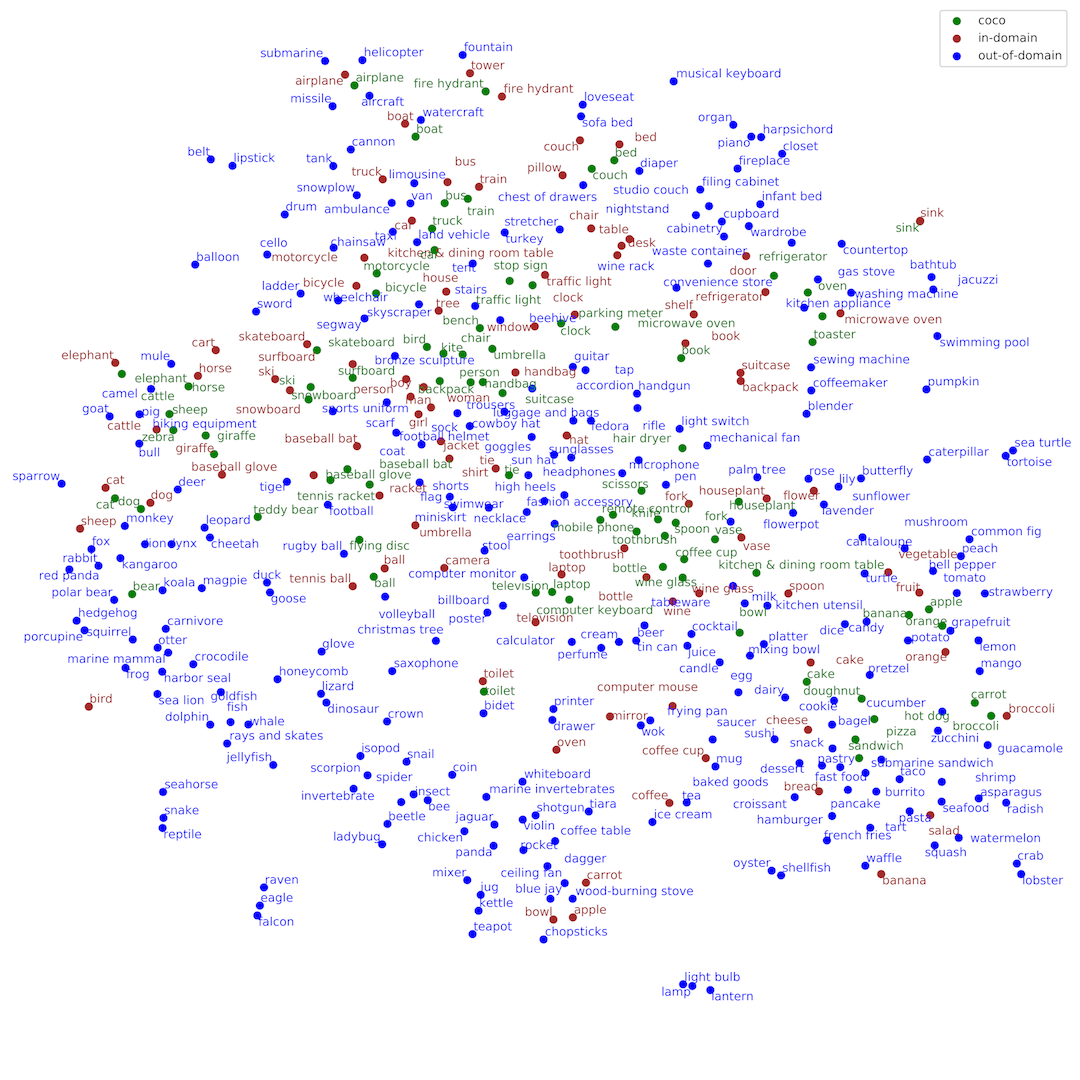}
\end{center}
\vspace{\captionReduceTop}
   \caption{T-SNE~\cite{maaten2008tsne} plot comparing the visual similarity between object classes in \coco, \nocapsin and \nocapsout splits of \nocaps. For each object class in a particular split, we extract bottom-up image features from the Faster-RCNN detector made publicly available by \cite{Anderson2017up-down} and mean pool them to form a 2048-dimensional vector. We further apply PCA on the feature vectors for all object classes and pick the first 128 principal components. Using these feature vectors of reduced dimension, we compute the exact form T-SNE with perplexity 30. We observe that: (a) \nocapsin shows high visual similarity to \coco -- green and brown points of same object class are close to each other. (b) Many \nocapsout classes are visually different from \nocapsin classes -- large clusters of blue, far away from green and brown. (c) \nocapsout also covers many visually similar concepts to \coco -- blue points filling the gaps between sparse clusters green/brown points.}
\vspace{\captionReduceBot}
\end{figure}

\clearpage

\subsection{T-SNE Visualization in Linguistic Feature Embedding Space}

\begin{figure}[h]
\begin{center}
\includegraphics[width=\linewidth]{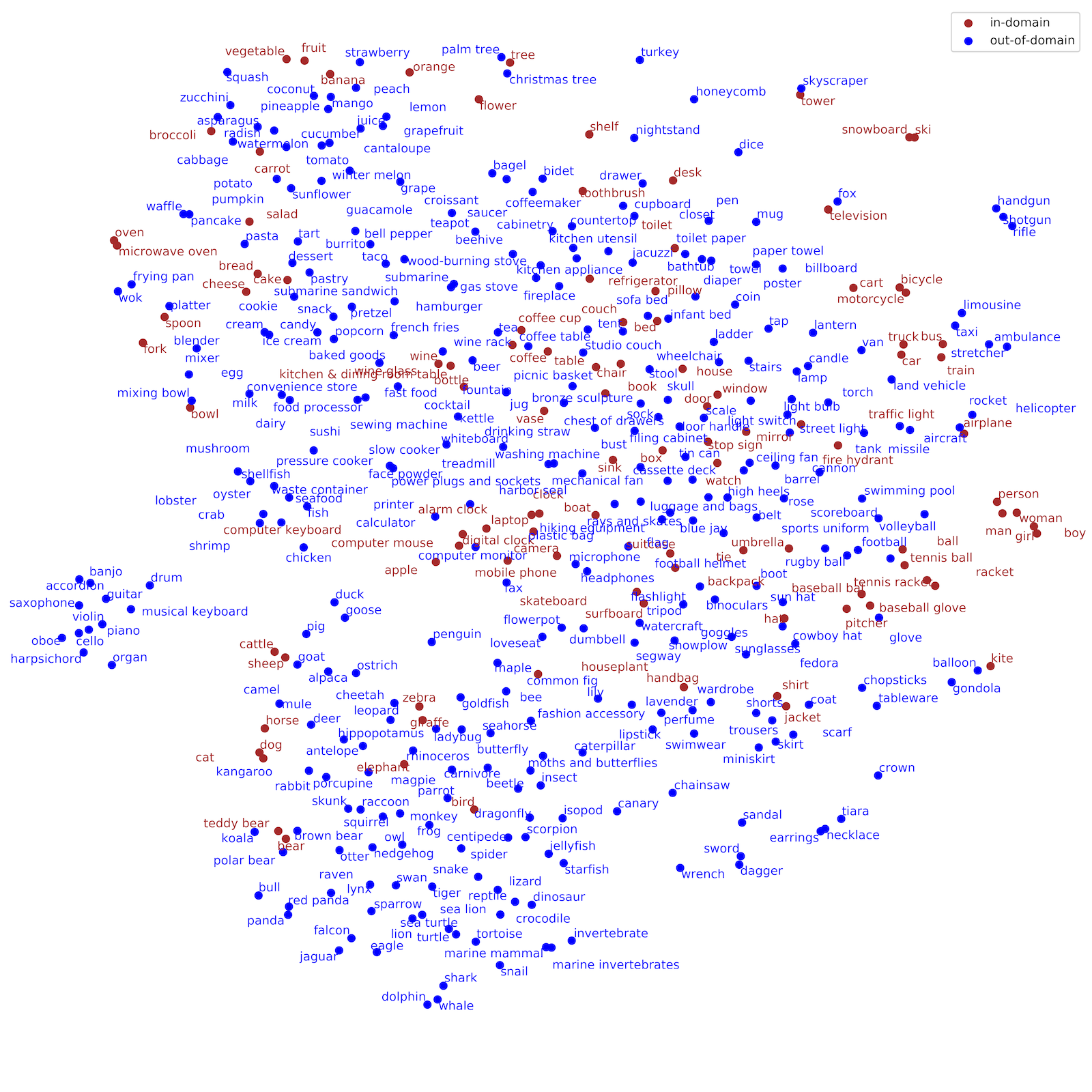}
\end{center}
\vspace{\captionReduceTop}
   \caption{T-SNE~\cite{maaten2008tsne} plot comparing the linguistic similarity between object classes in \nocapsin and \nocapsout splits of \nocaps. For each object class in a particular split, we obtain 300-dimensional GloVe~\cite{pennington2014glove}. We further apply PCA on these GloVe vectors vectors for all object classes and pick the first 128 principal components. Using these feature vectors of reduced dimension, we compute the exact form T-SNE with perplexity 30. We observe that: (a) Many \nocapsout classes are linguistically different from \nocapsin classes -- large clusters of blue points far away from brown points. (b) \nocapsout also covers many linguistically similar, fine-grained classes not present in \nocapsin -- blue points filling gaps in sparse clusters of brown points.}
\vspace{\captionReduceBot}
\end{figure}

\clearpage

\subsection{Linguistic Similarity to COCO}

Overall, our collection methodology closely follows \coco. However, we do introduce keyword priming to the collection interface (refer Figure \ref{fig:interface}) which has the potential to introduce some linguistic differences between \nocaps and \coco. To quantitatively assess linguistic differences between the two datasets, we review the performance of \coco-trained models on the \nocaps validation set while controlling for visual similarity to \coco. As a proxy for visual similarity to \coco, we use the average cosine distance in FC7 CNN feature space between each \nocaps image and the 10 closest \coco images. 

As illustrated in Table \ref{tab:sim}, the baseline \updown model (trained using \coco) exceeds human performance on the decile of \nocaps images which are most similar to \coco images (decile=1, avg. cosine distance=0.15), consistent with the trends seen in the \coco dataset. This suggests that the linguistic structure of \coco and \nocaps captions is extremely similar. As the \nocaps images become visually more distinct from \coco images, the performance of \updown drops consistently. This suggests that no linguistic variations have been introduced between \coco and \nocaps due to priming and the degradation in the performance is due to visual differences. Similar trends are observed for our best model (\updown + ELMo + CBS) although the performance degradation with increasing visual dissimilarity to \coco is much less.


\vspace{5mm}
\begin{table}[h]
    \begin{center}
    \setlength\tabcolsep{7.0pt}
    \renewcommand{\arraystretch}{1}
    \begin{tabularx}{\textwidth}{lccccccccccc}
    \toprule
        & \multicolumn{11}{c}{\textbf{\nocaps test CIDEr scores}} \\ 
        \cmidrule{2-12}
        Decile & 1 & 2 & 3 & 4 & 5 & 6 & 7 & 8 & 9 & 10 & Overall\\
        Avg Cosine Dist from COCO & 0.15 & 0.18 & 0.20 & 0.21 & 0.23 & 0.24 & 0.25 & 0.27 & 0.30 & 0.35 & \\
        \midrule
        \updown & \textbf{82.6} & 72.6 & 63.9 & 61.1 & 55.9 & 55.0 & 50.7 & 48.5 & 39.2 & 28.7 & 54.5 \\
        \updown + ELMo + CBS & \textbf{81.8} & 77.3 & 75.4 & 72.8 & 77.1 & 78.2 & 72.3 & 71.7 & 70.6 & 65.1 & 73.1 \\
        Human & 77.8 & \textbf{78.0} & \textbf{82.4} & \textbf{84.0} & \textbf{86.2} & \textbf{88.8} & \textbf{89.4} & \textbf{91.2} & \textbf{97.3} & \textbf{95.6} & \textbf{85.3}
 \\
     \bottomrule
    \end{tabularx}
    \end{center}
    \caption{CIDEr scores on \nocaps test deciles split by visual similarity to \coco (using CNN features). Our models exceed human performance on the decile of \nocaps images that are most visually similar to \coco. This suggests that after controlling for visual variations the linguistic structure of \coco and \nocaps captions is highly similar.}
    \label{tab:sim}
\end{table}

\clearpage

\vspace{\sectionReduceTop}
\section{Additional Implementation Details for Baseline Models} \label{sec:baseline_additional}
\vspace{\sectionReduceBot}

\subsection{Neural Baby Talk (NBT)}
\vspace{\subsectionReduceBot}

In this section, we describe our modifications to the original authors' implementation of Neural Baby Talk (NBT)~\cite{Lu2018Neural} to enable the model to produce captions for images containing novel objects present in \nocaps.

\noindent \textbf{Grounding Regions for Visual Words}

Given an image, NBT leverages an object detector to obtain a set of candidate image region proposals, and further produces a caption template, with slots explicitly tied to specific image regions. In order to accurately caption \nocaps images, the object detector providing candidate region proposals must be able to detect the object classes present in \nocaps (and broadly, \openimages). Hence, we use a Faster-RCNN \cite{faster_rcnn} model pre-trained using \openimages V4 \cite{openimages} (referred as \oidet henceforth), to obtain candidate region proposals as described in Section 4 of the main paper. This model can detect 601 object classes of \openimages, which includes the novel object classes of \nocaps. In contrast, the authors' implementation uses a Faster-RCNN trained using \coco.

For every image in \coco train 2017 split, we extract image region proposals after the second stage of detection, with an IoU threshold of 0.5 to avoid highly overlapping region proposals, and a class detection confidence threshold of 0.5 to reduce false positive detections. This results in number of region proposals per image varies up to a maximum of 18.

\noindent \textbf{Bottom-Up Visual Features}

The language model in NBT (Refer Figure 4 in \cite{Lu2018Neural}) has two separate attention layers, and takes visual features as input in three different manners:

\begin{compactitem}[\hspace{1pt}-]
\item The first attention layer learns an attention distribution over region features, extracted using ResNet-101 + RoI Align layer.
\item The second attention layer learns an attention distribution over spatial CNN features from the last convolutional layer of ResNet-101 (7 x 7 grid, 2048 channels).
\item The word embedding input is concatenated with FC7 features from ResNet-101 at every time-step.
\end{compactitem}

All the three listed visual features are extracted using ResNet-101, with the first being specific to visual words, while the second and third provide the holistic context of the image. We replace the ResNet-101 feature extractor with the publicly available Faster-RCNN model pre-trained using \visualgenome (referred as \vgdet henceforth), same as \cite{Anderson2017up-down}. Given a set of candidate region proposals obtained from \oidet, we extract 2048-dimensional bottom-up features using the \vgdet and use them as input to first attention layer (and also for input to the Pointer Network). For input to the second attention layer, we extract top-36 bottom-up features (class agnostic) using the \vgdet. Similarly, we perform mean-pooling of these 36 features for input to the language model at every time-step.

\noindent \textbf{Fine-grained Class Mapping}

NBT fills the slots in each caption template using words corresponding to the object classes detected in the corresponding image regions. However, object classes are coarse labels (e.g. `cake'), whereas captions typically refer entities in a fine-grained fashion (e.g. `cheesecake', `cupcake', `coffeecake' etc.). To account for these linguistic variations, NBT predicts a fine-grained class for each object class using a separate MLP classifier. To determine the output vocabulary for this fine-grained classifier we extend the fine-grained class mapping used for \coco (Refer Table 5 in \cite{Lu2018Neural}), adding \openimages object classes. Several fine-grained classes in original mapping are already present in \openimages (e.g. `man', `woman' -- fine-grained classes of `person'), we drop them as fine-grained classes from original mapping and retain them as \openimages object classes.

\noindent \textbf{Visual Word Prediction Criterion}

In order to ensure correctness in visual grounding, the authors' implementation uses three criteria to decide whether a particular region proposal should be tied with a "slot" in the caption template. At any time during decoding, when the Pointer Network attends to a visual feature (instead of the visual sentinel), the corresponding region proposal is tied with the "slot" if:

\begin{compactitem}[\hspace{1pt}-]
\item The class prediction threshold of this region proposal is higher than 0.5.
\item The IoU of this region proposal with at least one of the ground truth bounding boxes is greater than 0.5.
\item The predicted class is same as the object class of ground truth bounding box having highest IoU with this region proposal.
\end{compactitem}

We drop the third criterion, as the \oidet can predict several fine-grained classes in context of \coco, such as `man' and `woman' (while the ground truth object class would be `person'). Keeping the third criterion intact in \nocaps setting would suppress such region proposals, and result in lesser visual grounding, which is not desirable for NBT. Relaxation of this criterion might introduce false positives from detection in the caption but prevents reduction in visual grounding.

We encourage the reader to refer the authors' implementation for further details. We will release code for our modifications.

\subsection{Constrained Beam Search (CBS)}

\noindent \textbf{Determining Constraints}

When using constrained beam search (CBS)~\cite{cbs2017}, we decoded the model in question while forcing the generated caption to include words corresponding to object classes detected in the image. For object detection, we use the same Faster-RCNN \cite{faster_rcnn} model pre-trained using \openimages V4 \cite{openimages} (\oidet) that is used in conjunction with NBT. However, not all detected object classes are used as constraints. We perform constraint filtering by removing the 39 object classes listed in Table \ref{remove_class} from the constraint set, as these classes are either object parts, or classes that we consider to be either too rare or too broad. We also suppress highly overlapping objects as described in Section 4 of the main paper.

\begin{table}[ht]
\centering
\begin{tabular}{|c|c|}
\midrule
\textbf{Parts}             & \textbf{Too Rare or Too Broad}      \\ \midrule
Human Eye                  & Clothing           \\ 
Human Head                 & Footwear           \\ 
Human Face                 & Fashion Accessory  \\ 
Human Mouth                & Sports Equipment   \\ 
Human Ear                  & Hiking Equipment   \\
Human Nose                 & Mammal             \\ 
Human Hair                 & Personal Care      \\ 
Human Hand                 & Bathroom Accessory \\ 
Human Foot                 & Plumbing Fixture   \\ 
Human Arm                  & Tree               \\ 
Human Leg                  & Building           \\ 
Human Beard                & Plant              \\ 
Human Body                 & Land Vehicle       \\ 
Vehicle Registration Plate & Person             \\ 
Wheel                      & Man                \\
Seat Belt                  & Woman              \\ 
Tire                       & Boy                \\ 
Bicycle Wheel              & Girl               \\ 
Auto Part                  &                    \\ 
Door Handle                &                    \\ 
Skull                      &                    \\ 
\midrule
\end{tabular}
\caption{Blacklisted object class names for constraint filtering (CBS) and visual word prediction (NBT).}
\label{remove_class}
\end{table}

To quantify the impact of this simple constraint filtering heuristic, in Table \ref{tab:result-filter-objects} we report the results of the following ablation studies:
\begin{compactitem}[\hspace{1pt}-]
\item Using all the object classes for constraints (w/o class),
\item Using overlapping objects for constraints (w/o overlap), and
\item Using no filtering heuristic at all (w/o both).
\end{compactitem}
\noindent 
Note that in all cases we rank objects based on confident score for detected objects and pick the top-3 as the constraints. We report results for three models, the baseline model (\updown), the baseline model using Glove~\cite{pennington2014glove} and dependency-based~\cite{levy2014dependency} word embeddings (\updown + GD) and our ELMo-based model (\updown + ELMo +CBS). Table \ref{tab:result-filter-objects} shows that removing the above 39 classes significantly improves the performance of constrained beam search and removing overlapping objects can also slightly improve the performance. This conclusion is consistent across the three models.

\begin{table}[ht]
\begin{center}
\footnotesize
\setlength\tabcolsep{3pt}
\renewcommand{\arraystretch}{1.2}
\begin{tabularx}{\textwidth}{c@{~~}lYYYYYYYYY}
\toprule
& & \multicolumn{2}{c}{In-Domain} & \multicolumn{2}{c}{Near-Domain} & \multicolumn{2}{c}{Out-of-Domain}  & \multicolumn{2}{c}{Overall} \\
& & CIDEr & SPICE & CIDEr & SPICE & CIDEr & SPICE & CIDEr & SPICE \\ \midrule

\updown + CBS w/o both & & 73.4 & 11.2 & 68.0 & 10.9 & 65.2 & 9.8 & 68.2 & 10.7  \\
\updown + CBS w/o class &  & 72.8 & 11.2 & 68.6 & 10.9 & 65.5 & 9.7 & 68.6 & 10.8  \\
\updown + CBS w/o overlap  & & 80.6 & 12.0 & 73.5 & 11.3 & 66.4 & 9.8 & 73.1 & 11.1  \\
\updown + CBS  & & 80.0 & 12.0 & 73.6 & 11.3 & 66.4 & 9.7 & 73.1 & 11.1  \\  
\midrule

\updown + GD + CBS w/o both & & 72.8 & 11.2 & 68.4 & 10.8 & 66.3 & 9.8 & 68.6 & 10.7  \\
\updown + GD + CBS w/o class & & 72.3 & 11.2 & 68.6 & 10.9 & 66.9 & 9.7 & 68.8 & 10.7  \\
\updown + GD + CBS w/o overlap & & 77.0 & 12.0 & 73.5 & 11.4 & 67.2 & 9.7 & 72.8 & 11.1  \\
\updown + GD + CBS & & 77.0 & 12.0 & 73.6 & 11.4 & 69.5 & 9.7 & 73.2 & 11.1  \\
\midrule

\updown + ELMo + CBS w/o both  & & 73.3 & 11.5 & 68.6 & 10.9 & 70.0 & 10.8 & 69.6 & 10.8  \\
\updown + ELMo + CBS w/o class & & 73.5 & 11.5 & 69.2 & 11.0 & 69.9 & 9.9 & 70.0 & 10.9  \\
\updown + ELMo + CBS w/o overlap & & 79.8 & 12.3 & 73.7 & 11.4 & 72.0 & 9.9 & 74.2 & 11.2  \\
\updown + ELMo + CBS & &  \textbf{79.3} & \textbf{12.4} & \textbf{73.8} & \textbf{11.4} & \textbf{71.7} & \textbf{9.9} & \textbf{74.3} & \textbf{11.2}  \\ 
\midrule
Human & & \textbf{83.3} & \textbf{13.9} & \textbf{85.5} & \textbf{14.3} & \textbf{91.4} & \textbf{13.7} & \textbf{87.1} & \textbf{14.1} \\ 
\bottomrule
\end{tabularx}
\end{center}
\vspace{\captionReduceTop}
\caption{We investigate the effect of different object filtering strategies in Constrained Beam Search and report the model performance in \nocaps val. We find that using both strategies with the ELMo model performs best.}
\vspace{\captionReduceBot}
\label{tab:result-filter-objects}
\end{table}

\noindent \textbf{Finite State Machine}

\begin{figure}[!ht] 
\centering 
\begin{tikzpicture}

\node[state, initial] (q0) {$q_0$};
\node[state, right of=q0] (q1) {$q_1$};
\node[state, accepting, right of=q1] (q2) {$q_2$};

\draw (q0) edge[loop above] node{$V - {a_1}$} (q0)
(q0) edge[bend left,above] node{$a_1$} (q1)
(q1) edge[bend left, below] node{$V - {a_2}$} (q0)
(q1) edge[above] node{$a_2$} (q2)
(q2) edge[loop above] node{$V$} (q2);

\end{tikzpicture}
\caption{FSM for a two-word phrase $\{a_1, a_2\}$ constraint}
\label{fig:two_word}
\end{figure}
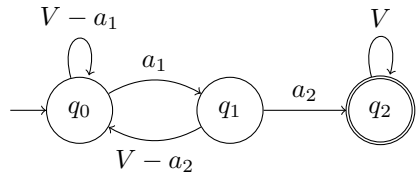

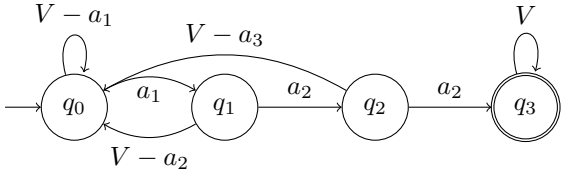
\begin{figure}[!ht] 
\centering 
\begin{tikzpicture}

\node[state, initial] (q0) {$q_0$};
\node[state, right of=q0] (q1) {$q_1$};
\node[state, right of=q1] (q2) {$q_2$};
\node[state, accepting, right of=q2] (q3) {$q_3$};

\draw (q0) edge[loop above] node{$V - {a_1}$} (q0)
(q0) edge[bend left,below] node{$a_1$} (q1)
(q1) edge[bend left, below] node{$V - {a_2}$} (q0)
(q2) edge[bend right, above] node{$V - {a_3}$} (q0)
(q1) edge[above] node{$a_2$} (q2)
(q2) edge[above] node{$a_2$} (q3)
(q3) edge[loop above] node{$V$} (q3);

\end{tikzpicture}
\caption{FSM for a three-word phrase $\{a_1, a_2, a_3\}$ constraint}
\label{fig:three_word}
\end{figure}

\begin{figure}[!ht] 
\centering 
\resizebox{.5\linewidth}{!}{
\begin{tikzpicture}
\tikzset{
    node distance=3.2cm
}
\node[state, initial] (q0) {$q_0$};
\node[state, right of=q0] (q1) {$q_1$};
\node[state, right of=q1] (q2) {$q_2$};
\node[state, accepting, right of=q2] (q3) {$q_3$};
\node[state, below of=q0] (q4) {$q_4$};
\node[state, accepting, below of=q1] (q5) {$q_5$};
\node[state, accepting, below of=q2] (q6) {$q_6$};
\node[state, accepting, below of=q3] (q7) {$q_7$};

\draw (q0) edge[loop above] node{} (q0)
(q1) edge[loop above] node{} (q1)
(q2) edge[loop above] node{} (q2)
(q3) edge[loop above] node{} (q3)
(q4) edge[loop below] node{} (q4)
(q5) edge[loop below] node{} (q5)
(q6) edge[loop below] node{} (q6)
(q7) edge[loop below] node{} (q7)
(q0) edge[above] node{$D_1$} (q1)
(q2) edge[above] node{$D_1$} (q3)
(q4) edge[above] node{$D_1$} (q5)
(q6) edge[above] node{$D_1$} (q7)
(q0) edge[bend left, above] node{$D_2$} (q2)
(q1) edge[bend left, above] node{$D_2$} (q3)
(q0) edge[above,left=0.2] node{$D_3$} (q4)
(q1) edge[above,left=0.2] node{$D_3$} (q5)
(q2) edge[above,left=0.2] node{$D_3$} (q6)
(q3) edge[above,left=0.2] node{$D_3$} (q7)
(q4) edge[bend right, below] node{$D_2$} (q6)
(q5) edge[bend right, below] node{$D_2$} (q7);

\end{tikzpicture}}
\caption{FSM for $D_1$, $D_2$, $D_3$ constraints. States $q_3$, $q_5$, $q_6$ and $q_7$ are desired states to satisfy at least two out of three constraints.}
\label{fig:eight_machine}
\end{figure}
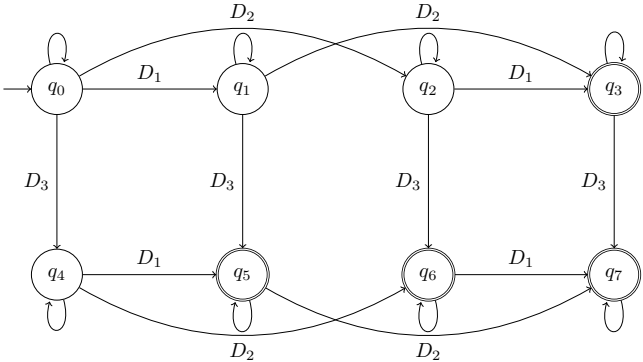

Constrained Beam Search implements constraints in the decoding process using a Finite State Machine (FSM). In all experiments we use a 24 state FSM. We use 8 states for standard three single word constraints $D_1$, $D_2$ and $D_3$. As shown in Figure~\ref{fig:eight_machine}, the outputs of this FSM are the captions that mention at least two constraints out of three. Each $D_i$ (i = 1,2,3) represents a set of alternative constraint words (e.g., bike, bikes). $D_i$ can also be multi-word expressions. Our FSM can dynamically support two-word or three-word phrases in $D_i$ by extending additional one states (see  Figure \ref{fig:two_word}) or two states (see Figure \ref{fig:three_word}) for two-word or three-word phrases respectively. Since $D_1$, $D_2$ and $D_3$ are all used 4 times in the base eight-state FSM, we need to allocate 4 states for a single two-word expression and 8 states for a single three-word expression.

\clearpage
\noindent \textbf{Integrating \updown Model with ELMo}

When using ELMo~\cite{N18-1202}, we use a dynamic representation of $w_c$, $\bar{h}_t^1$ and $\bar{h}_t^2$ as the input word embedding $w_{ELMo}^t$ for our caption model. $w_c$ is the character embedding of input words and $\bar{h}_t^i$ ($i=1,2$) is the hidden output of $i^{th}$ LSTM layer of ELMo. We combine them via:
\begin{align}
w_{ELMo}^t = \gamma_0 \cdot w_c +  \gamma_1 \cdot \bar{h}_t^1 + \gamma_2 \cdot \bar{h}_t^2
\label{elmo_cal}
\end{align}
where $\gamma_i$ (i=0, 1, 2) are three trainable scalars. When using $w_{ELMo}^t$ as the external word representation of other models, we fixed all the parameters of ELMo but $\gamma_i$ (i=0, 1, 2).

In addition, to handle unseen objects in training data, following~\cite{cbs2017}, we initialize the softmax layer matrix ($W_p$, $b_p$) using word embedding and keep this layer fixed during training. This allow our caption model to produce similar logits score for the words that share similar vectors and values in $W_p$ and $b_p$. We have: 
\begin{align}
W_p = W_{ELMo} \\
b_p = b_{ELMo}
\end{align}
where $W_{ELMo}$ and $b_{ELMo}$ is the softmax layer in original ELMo language model. To align the different dimension in softmax layer and LSTM hidden state, we add an additional fully connected layer with a non-linearity function $tanh$. We have:
\begin{gather}
    v_t = tanh(W_t h_t^2 + b_t) \\
P(y_t | y_{1:t-1}, I) = softmax(W_p v_t + b_p)
\end{gather}
where $W_t \in \mathbb{R}^{H \times E}$, $b_t \in \mathbb{R}^{E}$, $H$ is LSTM hidden dimension, $E$ is the word embedding dimension,  $W_p \in  \mathbb{R}^{E \times D}$, $b_p \in \mathbb{R}^{D}$ and $D$ is the vocabulary size. 

\noindent \textbf{Other details of using ELMo}

In our experiment, we use the full tensorflow checkpoint trained on 1 Billion Word Language Model Benchmark\footnote{\url{http://www.statmt.org/lm-benchmark/}} from official ELMo tensorflow implementation project\footnote{\url{https://github.com/allenai/bilm-tf/}}. 

When selecting vocabularies for our model, we first extract all words from COCO captions and open image object labels. We then extend the open image object labels to both singular and plural word forms. Finally, we remove all the words that are not in ELMo output vocabularies. This allow us to use ELMo LM prediction for each decoding step.

Our \updown + ELMo model is optimized by SGD~\cite{bottou2010large}. We conduct hyper-parameter tuning the model and choose the model based on its performance on \nocaps val.

\label{sec:outline}

\clearpage

\vspace{\sectionReduceTop}
\section{Example Model Predictions} \label{sec:prediction_examples}
\vspace{\sectionReduceBot}
\begin{figure*}[!ht]
	\hspace{-0.1in}\resizebox{0.99\textwidth}{!}{
	\renewcommand{\arraystretch}{1}
	\centering
	\small
	\begin{tabularx}{\textwidth}{lXXX}
		& \multicolumn{1}{c}{\nocapsin} & \multicolumn{1}{c}{\nocapsnear} & \multicolumn{1}{c}{\nocapsout} \\
		\multicolumn{1}{l}{\textbf{Method}}
        & \includegraphics[width=0.24\textwidth, clip]{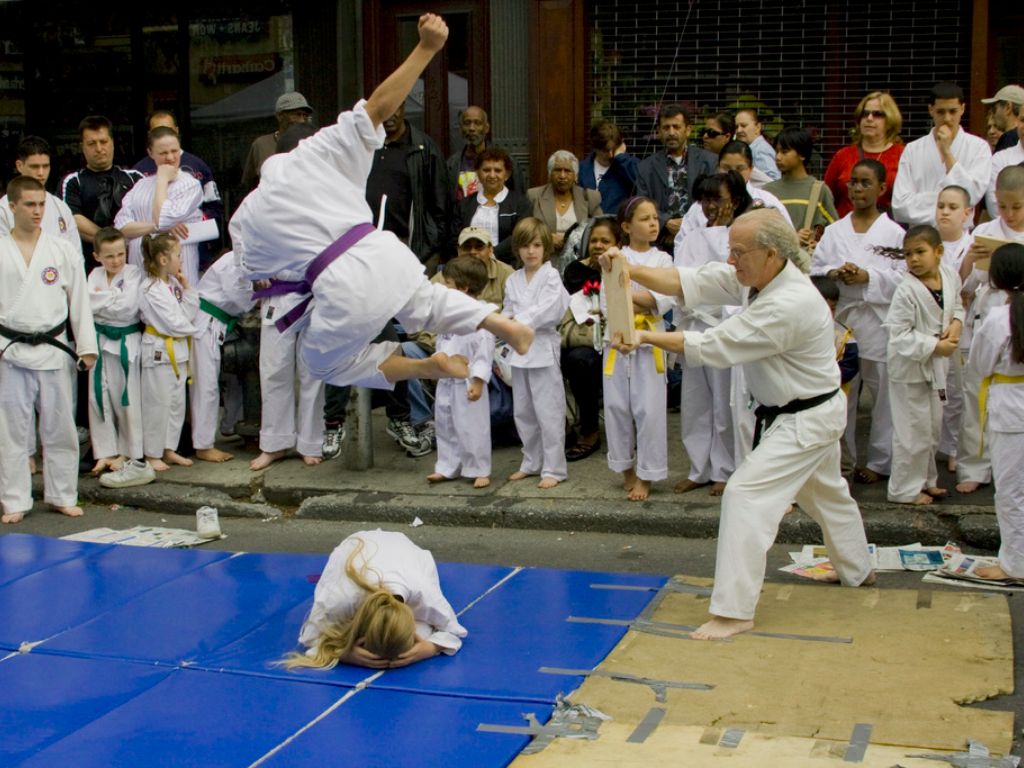}
        & \includegraphics[width=0.24\textwidth, clip]{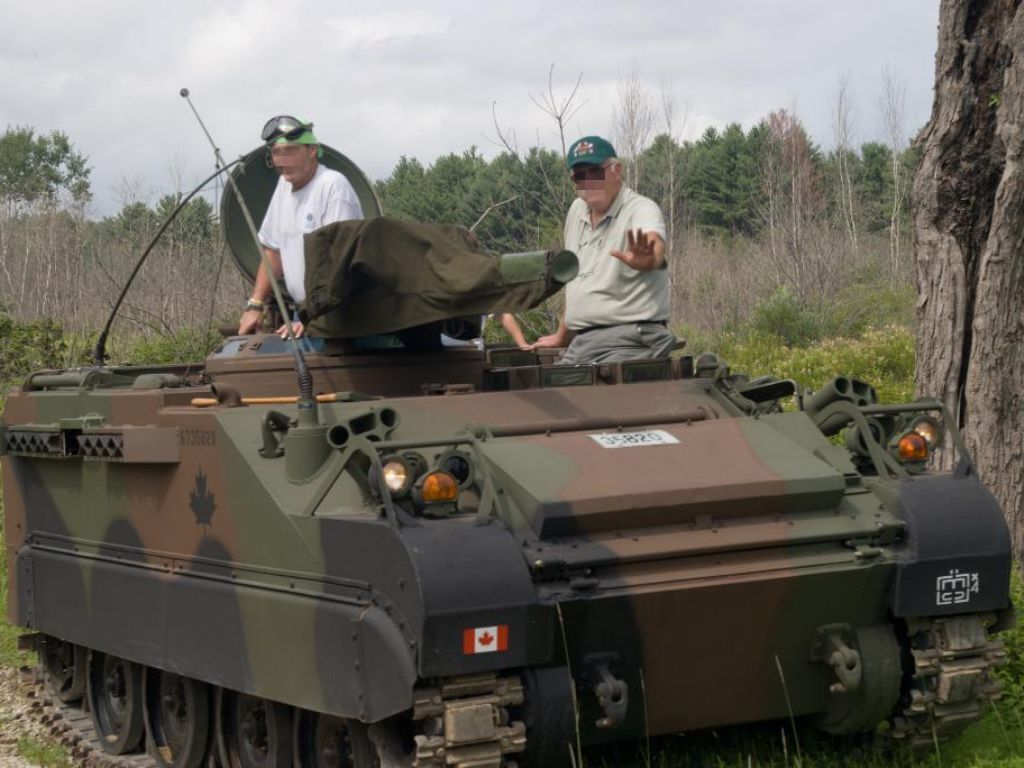}
        & \includegraphics[width=0.24\textwidth, clip]{figures/val/ba203a90803733a4.jpg} \\

		\cmidrule{1-4}
		\footnotesize \textbf{\updown}
		    & A man in a white shirt is playing baseball.
		    & A couple of men standing on top of a truck.
		    & A group of vases sitting on top of a table.\\
		\footnotesize \textbf{\updown + ELMo}
		    & A group of people standing around a blue table.
		    & A couple of men standing next to a truck.
		    & Two vases sitting next to each other on a table.\\
		\footnotesize \textbf{\updown + ELMo + CBS}
		    & A group of people standing near a blue table.
		    & A couple of men standing on top of a \cbs{tank}.
		    & A \cbs{teapot} sitting on top of a table next to a \cbs{vase}.\\
		\footnotesize \textbf{\updown + ELMo + CBS + GT}
		    & A group of people standing around a blue table.
		    & A couple of men standing on top of a \cbs{tank}.
		    & A couple of \cbs{kettle} \cbs{jugs} sitting next to each other.\\
		\midrule
		\footnotesize \textbf{NBT}
		    & A group of \nbt{men} standing in a field.
		    & A \nbt{man} standing on the back of a \nbt{tank}.
		    & A couple of \nbt{kettles} are sitting on a table.\\
		\footnotesize \textbf{NBT + CBS}
		    & A couple of \nbt{men} standing on a tennis court.
		    & A \nbt{man} standing on top of a \cbs{tank} with a truck.
		    & A close up of a \cbs{kettle} on a table.\\
		\footnotesize \textbf{NBT + CBS + GT}
		    & A group of \nbt{men} are standing in a field.
		    & A man standing on top of a \cbs{tank} \nbt{plant}.
		    & Two \cbs{kettles} and \nbt{teapot} \cbs{jugs} are sitting on a table.\\
		\midrule
		\footnotesize \textbf{Human}
		    & Two people in karate uniforms spar in front of a crowd.
		    & Two men sitting on a tank parked in the bush.
		    & Ceramic jugs are on display in a glass case.\\

		\addlinespace[0.25in]
		& \multicolumn{1}{c}{\nocapsin} & \multicolumn{1}{c}{\nocapsnear} & \multicolumn{1}{c}{\nocapsout} \\
		\multicolumn{1}{l}{\textbf{Method}}
		& \includegraphics[width=0.24\textwidth, clip]{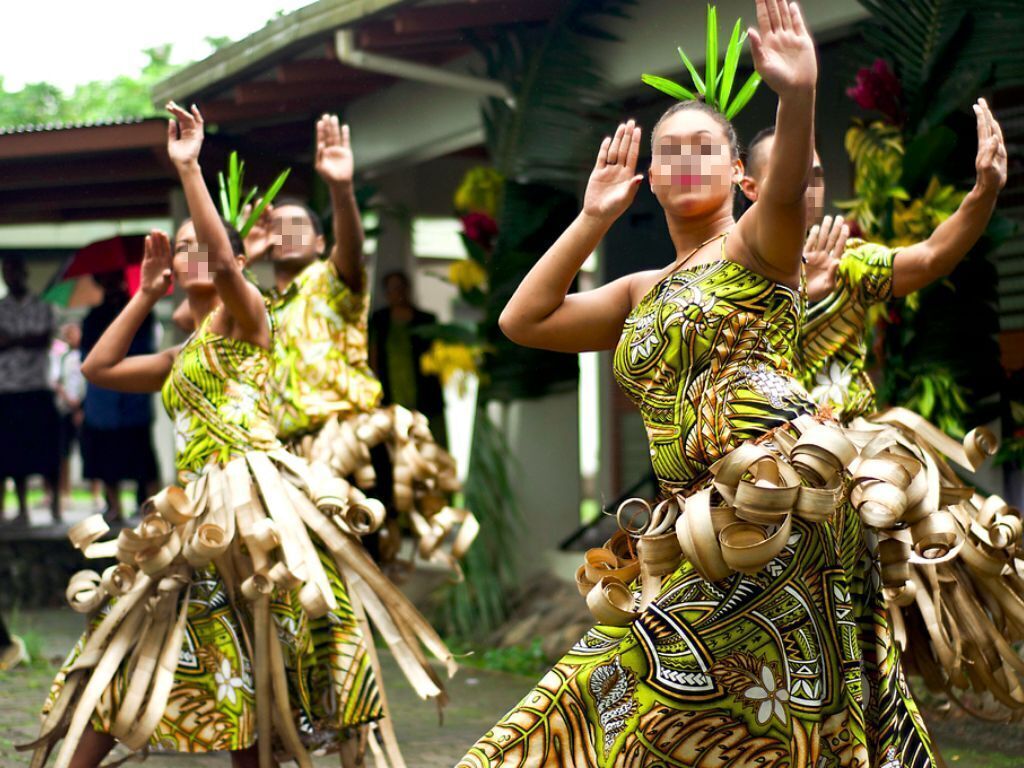}
        & \includegraphics[width=0.24\textwidth, clip]{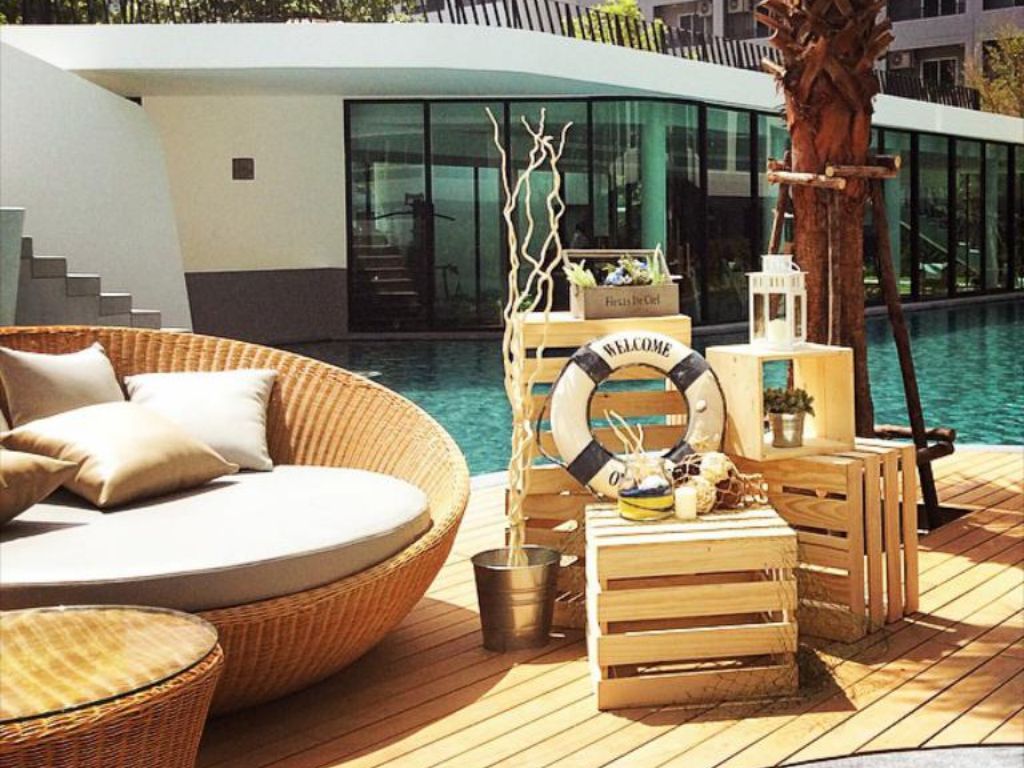}
        & \includegraphics[width=0.24\textwidth, clip]{figures/val/fa9c1a8fa79456a4.jpg} \\

		\cmidrule{1-4}
		\footnotesize \textbf{\updown}
		    & A woman riding a bike with a statue on her head.
		    & A couple of chairs sitting in front of a building.
		    & A bird sitting on the ground in the grass.\\
		\footnotesize \textbf{\updown + ELMo}
		    & There is a woman that is riding a bike.
		    & A room that has a lot of furniture in it.
		    & A dog laying on the ground next to a stuffed animal.\\
		\footnotesize \textbf{\updown + ELMo + CBS}
		    & There is a woman that is riding a bike.
		    & Two \cbs{pillows} and a table in the \cbs{house}.
		    & A dog laying on the ground next to a \cbs{tortoise}.\\
		\footnotesize \textbf{\updown + ELMo + CBS + GT}
		    & There is a woman that is riding a bike.
		    & Two \cbs{couches} and a table in a \cbs{house}.
		    & A dog laying on the ground next to a \cbs{tortoise}.\\
		\midrule
		\footnotesize \textbf{NBT}
		    & A man is riding a \nbt{clothing} on a bike.
		    & A table with a \nbt{couch} and a table.
		    & A \nbt{tortoise} is laying on top of the ground.\\
		\footnotesize \textbf{NBT + CBS}
		    & A woman is riding a \nbt{clothing} in the street.
		    & A couple of \cbs{pillows} on a wooden table in a \cbs{couch}.
		    & A \cbs{tortoise} that is sitting on the ground.\\
		\footnotesize \textbf{NBT + CBS + GT}
		    & A man is riding a \nbt{clothing} on a \nbt{person}.
		    & A \nbt{house} and a \cbs{studio couch} of \cbs{couches} in a room.
		    & A \cbs{tortoise} is laying on the ground in the grass.\\
		\midrule
		\footnotesize \textbf{Human}
            & People are performing in an open cultural dance.
            & On the deck of a pool is a couch and a display of a safety ring.
            & Three tortoises crawl on soil and wood chips in an enclosure.\\

	\end{tabularx}}
	\caption{Some challenging images from \nocaps and corresponding captions generated by existing approaches. The constraints given to the CBS are shown in \cbs{blue}. The visual words associated with NBT are shown in \nbt{red}.}
	\label{fig:pred_examples_1}
	\vspace{-10pt}
\end{figure*}

\begin{figure*}[!ht]
	\hspace{-0.1in}\resizebox{0.99\textwidth}{!}{
	\renewcommand{\arraystretch}{1}
	\centering
	\small
	\begin{tabularx}{\textwidth}{lXXX}
		& \multicolumn{1}{c}{\nocapsin} & \multicolumn{1}{c}{\nocapsnear} & \multicolumn{1}{c}{\nocapsout} \\
		\multicolumn{1}{l}{\textbf{Method}}
        & \includegraphics[width=0.24\textwidth, clip]{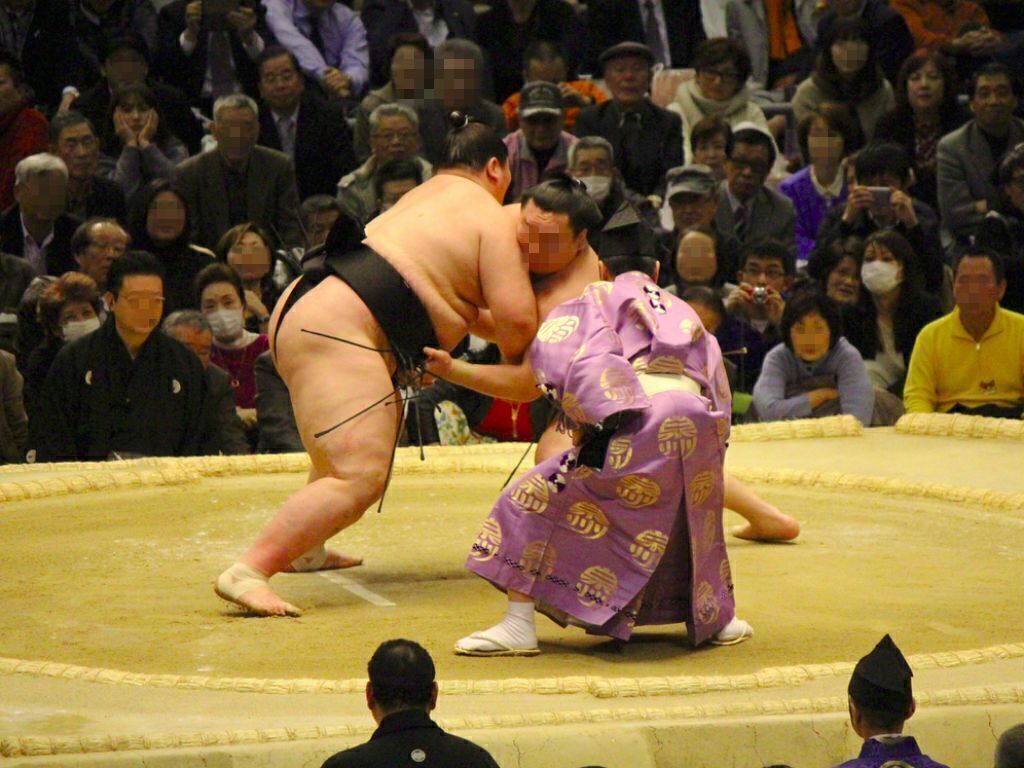}
        & \includegraphics[width=0.24\textwidth, clip]{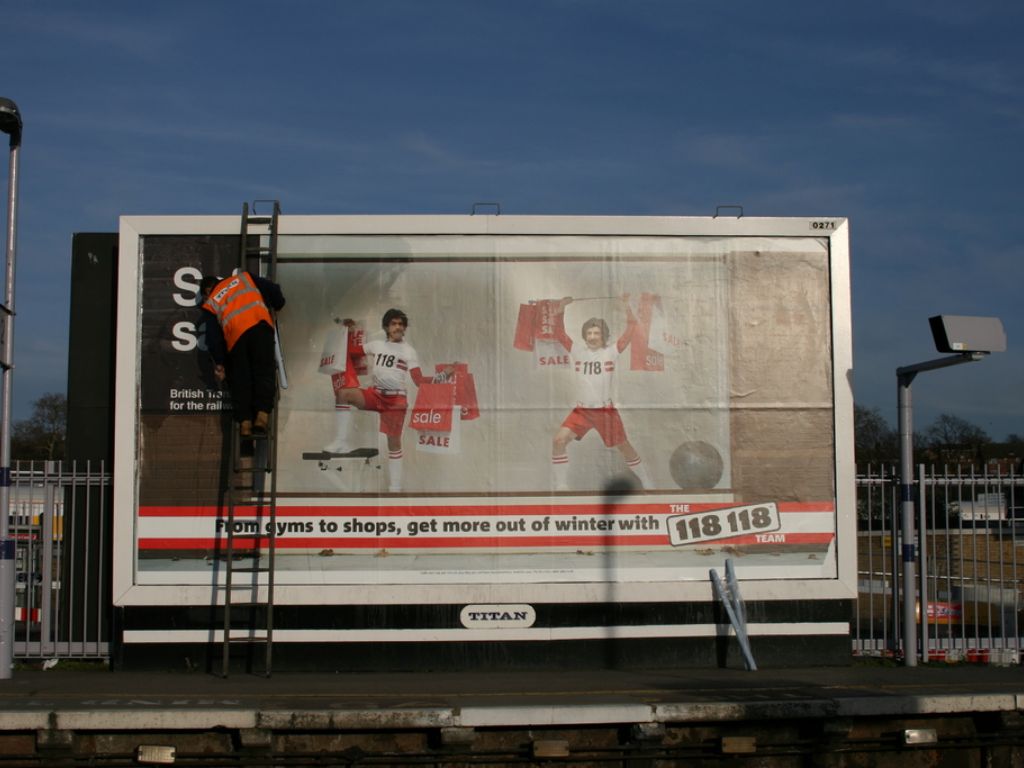}
        & \includegraphics[width=0.24\textwidth, clip]{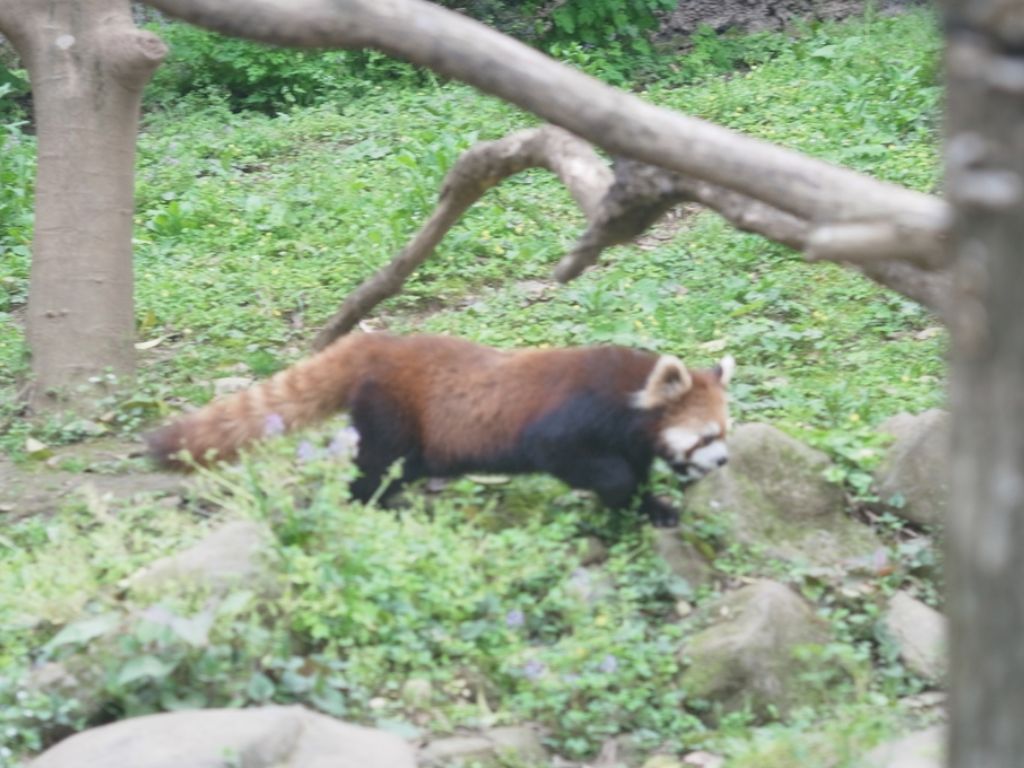}\\
        
		\cmidrule{1-4}
		\footnotesize \textbf{\updown}
		    & A group of people are playing a game.
		    & A large white sign on a city street.
		    & A bear laying in the grass near a tree.\\
		\footnotesize \textbf{\updown + ELMo}
		    & A woman in a pink dress is holding a child.
		    & A large white bus parked on the side of a road.
		    & A bear that is laying down in the grass.\\
		\footnotesize \textbf{\updown + ELMo + CBS}
		    & A woman in a pink dress is holding a child.
		    & A \cbs{billboard} that has a \cbs{street light} on it.
		    & A \cbs{red panda} is walking through the grass.\\
		\footnotesize \textbf{\updown + ELMo + CBS + GT}
		    & A woman in a pink dress is holding a child.
		    & A large white bus parked next to a \cbs{billboard}.
		    & A \cbs{red panda} is walking through the grass.\\
		\midrule
		\footnotesize \textbf{NBT}
		    & A group of \nbt{man} are standing in a field.
		    & A \nbt{billboard} sign on the side of a building.
		    & A brown \nbt{red panda} is laying on the grass.\\
		\footnotesize \textbf{NBT + CBS}
		    & A group of \nbt{man} are playing a baseball game.
		    & A picture of \cbs{billboard} sign on the \cbs{street light}.
		    & A \nbt{tree} and a brown \cbs{red panda} in a field.\\
		\footnotesize \textbf{NBT + CBS + GT}
		    & A group of \nbt{man} are standing on a field.
		    & A \cbs{billboard} sign on the side of a building.
		    & A brown \cbs{red panda} lying on top of a field.\\
		\midrule
		\footnotesize \textbf{Human}
		    & Two sumo wrestlers are wrestling while a crowd of men and women watch.
            & A man is standing on the ladder and working at the billboard.
            & The red panda trots across the forest floor.\\

		\addlinespace[0.25in]
		& \multicolumn{1}{c}{\nocapsin} & \multicolumn{1}{c}{\nocapsnear} & \multicolumn{1}{c}{\nocapsout} \\
		\multicolumn{1}{l}{\textbf{Method}}
	    & \includegraphics[width=0.24\textwidth, clip]{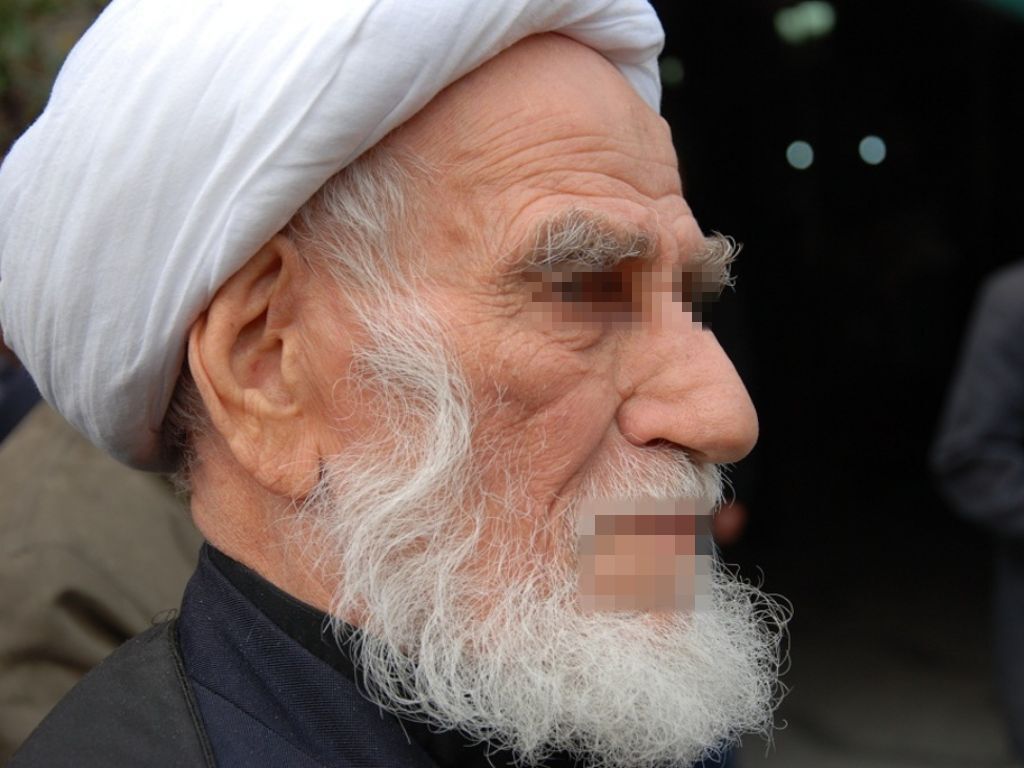}
        & \includegraphics[width=0.24\textwidth, clip]{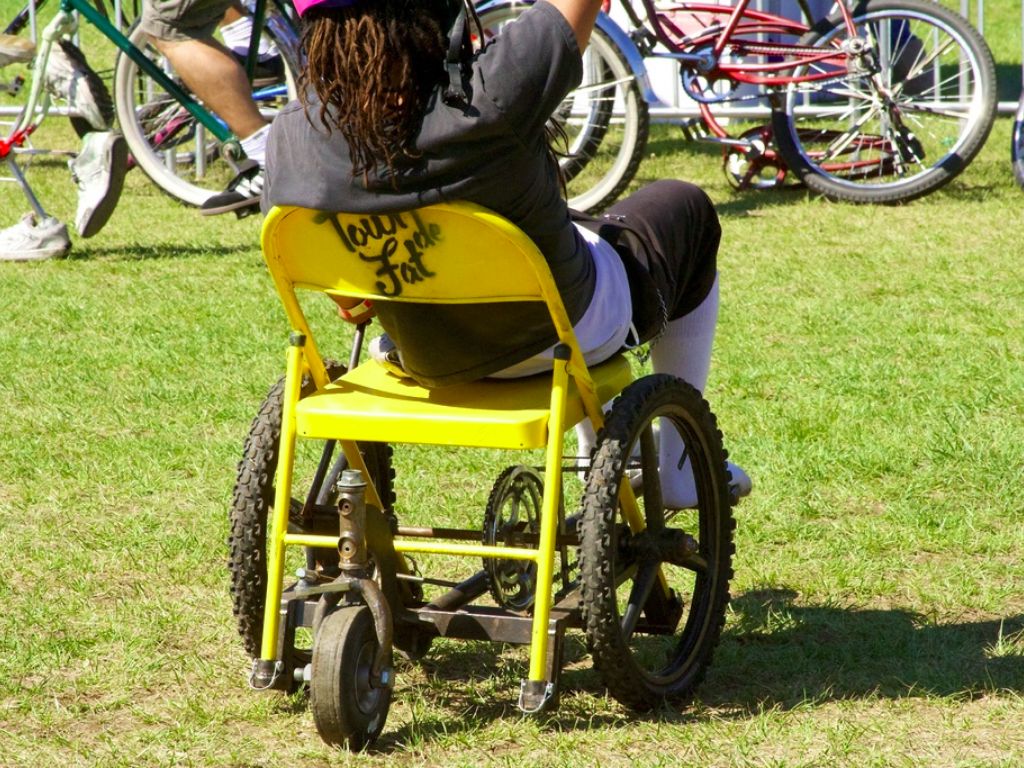}
        &  \includegraphics[width=0.24\textwidth, clip]{figures/val/08633d6cdda962f2.jpg}\\
        
		\cmidrule{1-4}
		\footnotesize \textbf{\updown}
		    & A woman wearing a white shirt and a white shirt.
		    & A person riding a yellow bike in the field.
		    & A close up of a cat looking at a bird.\\
		\footnotesize \textbf{\updown + ELMo}
		    & A woman wearing a white shirt and white shirt.
		    & A woman sitting on a yellow bike.
		    & A close up of a bird with its mouth open.\\
		\footnotesize \textbf{\updown + ELMo + CBS}
		    & A woman wearing a white \cbs{suit} and a white shirt.
		    & A person sitting on a bicycle with a \cbs{wheelchair}.
		    & A \cbs{sea lion} standing with its mouth open.\\
		\footnotesize \textbf{\updown + ELMo + CBS + GT}
		    & A woman wearing a white \cbs{suit} and a white shirt.
		    & A person sitting on a \cbs{bicycle} with a \cbs{wheelchair}.
		    & A \cbs{sea lion} standing next to a \cbs{harbor seal}.\\
		\midrule
		\footnotesize \textbf{NBT}
		    & A woman wearing a white shirt is wearing a hat.
		    & A man sitting on a \nbt{wheelchair} with a \nbt{bike}.
		    & A close up of a \nbt{sea lion} and \nbt{harbor seal} with its head.\\
		\footnotesize \textbf{NBT + CBS}
		    & A \cbs{suit} of woman wearing a white suit.
		    & A man sitting on a \cbs{wheelchair} and a \nbt{bike}.
		    & A close up of a \nbt{harbor seal} of a \cbs{sea lion}.\\
		\footnotesize \textbf{NBT + CBS + GT}
		    & A \cbs{suit} of woman wearing a white shirt.
		    & A \cbs{bicycle} sitting on a \cbs{wheelchair} with a \nbt{bike}.
		    & A close up of a \cbs{harbor seal} of a \cbs{sea lion}.\\
		\midrule
		\footnotesize \textbf{Human}
		    & The man has a wrap on his head and a white beard.
            & A person sitting in a yellow chair with wheels.
            & A brown and gray sea lion looking at the photographer. \\
	\end{tabularx}}
	\caption{Some challenging images from \nocaps and corresponding captions generated by existing approaches. The constraints given to the CBS are shown in \cbs{blue}. The visual words associated with NBT are shown in \nbt{red}.}
	\label{fig:pred_examples_1}
	\vspace{-10pt}
\end{figure*}

\vspace{5pt}

\end{document}